%% file: ijcai25.tex
\documentclass{article}
\usepackage{ijcai25}

\usepackage{times}
\usepackage{soul}
\usepackage{url}
\usepackage[hidelinks]{hyperref}
\usepackage[utf8]{inputenc}
\usepackage[small]{caption}
\usepackage{graphicx}
\usepackage{amsmath}
\usepackage{amsthm}
\usepackage{booktabs}
\usepackage{amsmath}
\usepackage{graphicx}
\usepackage{algorithm}
\usepackage[noend]{algpseudocode}

\usepackage{multirow}
\usepackage{xcolor}
\usepackage{tcolorbox}
\usepackage{adjustbox}
\usepackage{bbm}
\usepackage{wrapfig,lipsum}
\usepackage{xspace}
\usepackage{booktabs}       
\usepackage[normalem]{ulem}
\usepackage{makecell}
\usepackage{amssymb}
\usepackage{pifont}

\usepackage{cleveref}
\usepackage{float} 
\usepackage{url}

\newtheorem{proposition}{Proposition}[section]

\title{MMP-A*: Multimodal Perception Enhanced Incremental Heuristic Search on Path Planning\\
}

\author{
Minh Hieu Ha$^{1,2,*}$ \and
Khanh Ly Ta$^{2,*}$ \and
Hung Phan$^{1}$ \and
Tung Doan$^{1}$ \and
Tung Dao$^{1}$ \and
Dao Tran$^{3}$ \and
Huynh Thi Thanh Binh$^{1}$\\
\affiliations
$^1$Hanoi University of Science and Technology, Vietnam\\
$^2$Vingroup Big Data Research Center\\
$^3$FPT Software AI Center, Vietnam\\
}

\begin{document}

\maketitle
\renewcommand{\thefootnote}{*}
\footnotetext{Equal contribution.}
\renewcommand{\thefootnote}{\arabic{footnote}} 

\begin{abstract}
Autonomous path planning demands a synergy of global reasoning and geometric precision, particularly in complex or cluttered environments. While classical A* is favored for its optimality, it suffers from prohibitive computational and memory costs in large-scale scenarios. 
Recent efforts to mitigate these limitations by leveraging Large Language Models for waypoint guidance remain insufficient; operating solely on text-based reasoning without spatial grounding. These models frequently generate erroneous waypoints in topologically complex environments featuring dead-ends and lack the perceptual capacity to interpret ambiguous physical boundaries. Such inconsistencies necessitate costly corrective expansions, ultimately undermining the intended computational efficiency.
We introduce \textbf{MMP-A*}, a multimodal framework that synergizes the spatial grounding capabilities of Vision–Language Models with a novel adaptive decay mechanism. By anchoring high-level reasoning in physical geometry, our approach generates coherent waypoint guidance that effectively overcomes the blindness of text-only planners. The integrated decay mechanism dynamically regulates the influence of uncertain waypoints within the heuristic, ensuring geometric validity while substantially reducing memory overhead. To assess robustness, we evaluate the framework within challenging scenarios featuring severe clutter and topological complexity. Empirical results demonstrate that MMP-A* yields near-optimal trajectories with significantly reduced operational costs, confirming its potential as a robust, perception-grounded paradigm for autonomous navigation.

Code is available at: \url{https://github.com/langkhachhoha/MMP-ASTAR}
\end{abstract}

\input{sections/1_introduction}
\input{figures/sample}

\input{sections/2_relatedwork}

\input{figures/algorithm}

\input{sections/3_methodology}

\input{sections/4_experiments}

\input{sections/5_conclusion}

\bibliographystyle{named}
\bibliography{ijcai25}
\clearpage
\appendix

\input{sections/6_appendix}

\end{document}

%% file: sections/1_introduction.tex
\section{Introduction}
Path planning is a core problem in robotics and autonomous navigation, requiring the computation of an optimal or near-optimal route from start to goal while avoiding obstacles, and serving as a foundational capability for mobile robots, autonomous vehicles, industrial automation, and virtual environments where navigation efficiency directly impacts safety and scalability \cite{hart1968formal,abd2015comprehensive,gonzalez2015review}. Classical search-based methods, particularly A* and its derivatives, are widely utilized for their theoretical guarantees of completeness and optimality under admissible heuristics \cite{koenig2004lifelong,karaman2011sampling}. However, increasing scale and clutter trigger exponential computational costs in these algorithms. While specialized A* variants offer optimizations, they often rely on specific environmental priors, rendering them brittle and unable to generalize to the arbitrary, unstructured geometries of real-world settings.

Large Language Models (LLMs) have increasingly been applied across a wide range of domains, including robotics, where they support tasks such as high-level task planning and action selection \cite{tariq2025robust,doma2024llm,joublin2024copal}. 
In particular, LLM-A* \cite{meng2024llm} frameworks employ LLMs to generate intermediate waypoints that guide the search process toward semantically meaningful regions, significantly reducing expansion cost compared to vanilla A*. 
While this integration introduces promising global reasoning, the text-only modality of LLMs cannot encode fine-grained geometric or topological structures \cite{caglar2024can,wei2025plangenllms,cao2025large}. Consequently, in complex environments with dense or irregular barriers, LLM-generated waypoints often become redundant, misplaced, or geometrically infeasible, leading to unstable heuristic evaluations and degraded performance, sometimes worse than A* itself. 
Vision-Language Models (VLMs) partially address this issue by incorporating spatial grounding from visual inputs, enabling recognition of navigable areas and geometric relationships \cite{ye2025vlm}. However, inherent limitations in long-horizon planning render both LLMs and VLMs ineffective as standalone end-to-end motion planners \cite{aghzal2023can,aghzal2024look,yang2025guiding}. This constraint necessitates a paradigm shift: rather than serving as autonomous controllers, these models may be more effectively utilized as supportive modules within a structured planning framework. The complementary strengths of LLMs and VLMs, balancing high-level semantic reasoning with precise geometric grounding, thus motivate a multimodal integration designed to unify abstract cognition with physical spatial constraints.

To this end, we propose \textbf{MMP-A*}, which improves upon LLM-A* to overcome its limitations by integrating the global reasoning of LLMs, the spatial perception of VLMs, and the deterministic guarantees of A* into a unified framework. MMP-A* operates through a three-stage pipeline: (1) an LLM first generates a coarse waypoint sequence reflecting high-level navigation intent; (2) a VLM refines these waypoints by visually analyzing the environment, pruning redundant or infeasible checkpoints; and (3) an adaptive decay mechanism dynamically regulates the influence of VLM-validated waypoints in the heuristic function, attenuating their effect as uncertainty grows during search. This design prevents overreliance on stale or erroneous guidance while preserving the computational efficiency of waypoint-based exploration. Through this synergy of perception and reasoning, MMP-A* achieves a robust balance between geometric fidelity, efficiency, and scalability, especially in high-complexity maps characterized by dense barriers and intricate obstacle configurations.
Our main contributions are summarized as follows:
\begin{itemize}
    \item We propose \textbf{MMP-A*}, a multimodal framework that seamlessly unifies LLM-based reasoning, VLM-driven spatial grounding, and adaptive heuristic modulation, thereby enabling computationally efficient and geometrically reliable path generation.
    \item We introduce an adaptive decay mechanism that dynamically modulates waypoint influence, maintaining balanced exploration–exploitation and preventing bias toward uncertain guidance.
    \item We conduct rigorous evaluations within high-complexity environments characterized by dense obstacles and intricate layouts. Our findings show that MMP-A* delivers significantly superior memory efficiency without compromising path optimality.
\end{itemize}

%% file: figures/sample.tex
\begin{figure*}[!tb]
	\centering
	\includegraphics[width=0.8\linewidth] {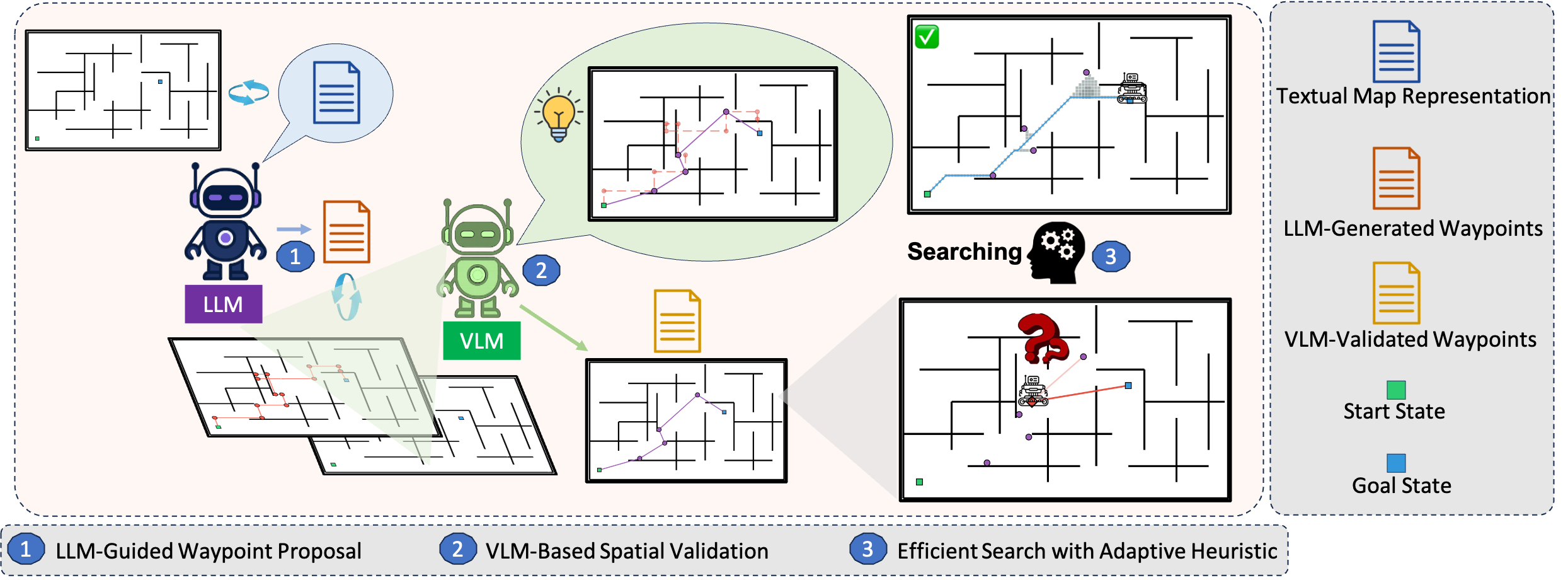}
	\caption{
		\textbf{\small Overall framework of MMP-A*:} The proposed planner operates in three stages: (1) the LLM analyzes the map and generates coarse waypoint suggestions; (2) the VLM refines these by visually filtering redundant or invalid checkpoints; and (3) the refined waypoints guide the A* search through an adaptive fading-checkpoint heuristic, producing valid and efficient paths in complex environments.
		\label{fig:schema}}
\end{figure*}

%% file: sections/2_relatedwork.tex
\section{Related Work}

\paragraph{Traditional Algorithms in Path Planning.}
Pathfinding has long been a fundamental problem in artificial intelligence, robotics, and computer graphics. The A* algorithm \cite{hart1968a} remains foundational for combining heuristic estimation with optimality guarantees, inspiring numerous extensions to improve efficiency and adaptability. 
Notable A* variants typically align with two distinct operational paradigms. In known environments, strategies such as IDA* \cite{korf1985depth}, RTA*/LRTA* \cite{korf1990real}, and SMA* \cite{russell1992memory} address the equilibrium between computational cost, memory constraints, and responsiveness. In contrast, dynamic environments necessitate algorithms like D* \cite{stentz1994optimal}, LPA* \cite{koenig2004lifelong} or Bug1 \cite{lumelsky1987path}  which are explicitly engineered to facilitate efficient incremental replanning.
To enhance scalability, hierarchical and structure-aware approaches like HPA* \cite{botea2004near} and JPS \cite{harabor2011online} further reduce search overhead, underscoring the enduring influence of heuristic search in modern path planning. In this context, our work specifically targets large-scale, known environments, aiming to address the scalability challenges posed by high-complexity static maps.

\paragraph{Large Language Models in Path Planning.}
LLMs have recently demonstrated impressive reasoning and generalization abilities across natural language processing and decision-making domains \cite{naveed2023comprehensive,chang2024survey,ha2025pareto}. Their capacity to decompose instructions and infer latent goals has motivated applications in high-level planning and embodied control \cite{erdogan2025plan,yang2025planllm,zhao2023large}.
To address the computational complexity of path planning, works like \cite{meng2024llm} and \cite{tariq2025robust} utilize LLMs to generate high-level waypoints, effectively pruning the search space and guiding geometric planners toward optimal solutions in large-scale environments. Conversely, some approaches \cite{wang2024llmˆ,doma2024llm,oelerich2024language} focus on trajectory feasibility, employing LLMs to translate semantic instructions into dynamic cost map updates or task-space constraints, ensuring the generation of safe, collision-free paths under complex restrictions.
However, LLMs still exhibit notable limitations in long-horizon planning and spatial reasoning, especially in tasks requiring continuous geometric understanding \cite{aghzal2023can,ilharco2020probing,patel2021mapping,abdou2021can}. 
To address this, we propose a hybrid framework that boosts high-level planning by coordinating LLMs with VLMs to anchor reasoning in physical space, coupled with a low-level A* planner to deliver mathematically accurate pathfinding results.

\paragraph{Vision Language Models in Path Planning.}
VLMs have emerged as a promising solution to these spatial reasoning limitations by jointly processing visual and linguistic information \cite{zhang2024vision,han2025multimodal,xu2025llava}. Through visual grounding, VLMs directly perceive environmental geometry, allowing them to assess navigability and spatial patterns that remain opaque to text-only models.
Recent frameworks utilize VLMs to ground search in visual reality, employing techniques like semantically-biased sampling \cite{ye2025vlm} or visual affordance prompting \cite{chen2025affordances} to steer planners toward safe, goal-relevant regions. Beyond generation, research also explores using VLMs as zero-shot reward models \cite{aghzal2025evaluating}, evaluating trajectory compliance against complex constraints.
Despite their perceptual capabilities, VLMs struggle to translate pixel-level information into continuous geometric coordinates, making visual input alone inadequate for rigorous motion planning \cite{baghaei2025follow,colan2025assessing}.

%% file: figures/algorithm.tex
\begin{figure}[t]
    \centering
    \begin{minipage}{1\linewidth}
        \begin{algorithm}[H]
            \small
            \caption{MMP-A* Algorithm with Adaptive Decay}
            \label{alg:mmp-a-star}
            \begin{algorithmic}[1]
                \State \textbf{Require:} START state $s_0$, GOAL state $s_g$, OBSTACLE set $\mathcal{S}_{obs}$, heuristic function $h()$, cost function $g()$, $llm()$, $vlm()$, adaptive decay factor $\alpha \in (0,1)$
                
                \State $\mathcal{O}_\text{open}\gets\{s_0\}$, $\mathcal{C}_\text{close}\gets\{\}$, TARGET list $\mathcal{T}\gets vlm(llm(s_0, s_g, \mathcal{O}))$, TARGET state $t\gets\mathcal{T}_\text{start}$, $g(s_0)\gets 0$, $f(s_0)\gets h(s_0)$, $k \gets 0$
                
                \While{$\mathcal{O}_\text{open} \neq \emptyset$}
                    \State $s_a \gets \arg\min_{s \in \mathcal{O}_\text{open}} f(s)$
                    \If{$s_a = s_g$}
                        \State \textbf{return} reconstruct\_path($s_a$)
                    \EndIf
                    \State Remove $s_a$ from $\mathcal{O}_\text{open}$
                    \State Add $s_a$ to $\mathcal{C}_\text{close}$
                    \ForAll{neighbors $s_n$ of $s_a$}
                        \If{$s_n = t$ \textbf{and} $s_g \neq t$}
                            \State $t \gets \mathcal{T}_\text{next}$ \Comment{Move to next waypoint}
                            \State $k \gets k + 1$ \Comment{Increment decay step}
                            \State Update $f$-cost of states in $\mathcal{O}_\text{open}$
                        \EndIf
                        \If{$s_n \in \left(\mathcal{C}_\text{close} \cup \mathcal{S}_{obs}\right)$}
                            \State \textbf{continue}
                        \EndIf
                        \State Tentative cost $g_\text{tent} \gets g(s_a) + \text{cost}(s_a, s_n)$
                        \If{$s_n \notin \mathcal{O}_\text{open}$ \textbf{or} $g_\text{tent} < g(s_n)$}
                            \State Update path to $s_n$ to go through $s_a$
                            \State $g(s_n) \gets g_\text{tent}$
                            \State $f(s_n) \gets g(s_n) + h_{A^*}(s_n) + \alpha^{k} \cdot \text{cost}(t, s_n)$
                            \If{$s_n \notin \mathcal{O}_\text{open}$}
                                \State Add $s_n$ to $\mathcal{O}_\text{open}$
                            \EndIf
                        \EndIf
                    \EndFor
                \EndWhile
                \State \textbf{return} \textbf{failure}
            \end{algorithmic}
        \end{algorithm}
    \end{minipage}
\end{figure}

%% file: sections/3_methodology.tex
\section{Methodology}
\subsection{Problem Formulation}
We formulate the navigation task as a path planning problem on a 2D grid map $\mathcal{M}$. 
The map is partitioned into free space $\mathcal{S}_{free}$ and an obstacle set $\mathcal{S}_{obs}$, which consists of impassable barriers. For irregular or non-grid-aligned obstacles, we simplify their representation using axis-aligned orthogonal segments centered at their centroids to facilitate LLM-based symbolic reasoning (see Appendix \ref{represntation barier} for more details).

The robot is initialized at a starting state $s_0$ and aims to reach a goal state $s_g$. We adopt an 8-connected grid topology (Moore neighborhood), allowing the robot to transition to any adjacent cell $s' \in \mathcal{S}_{free}$. The transition cost is defined as the Euclidean distance: $1$ for cardinal and $\sqrt{2}$ for diagonal movements. The objective is to generate a sequence of states $P = \{s_0, s_1, \dots, s_g\}$ such that every state in $P$ avoids $\mathcal{S}_{obs}$, ensuring a collision-free trajectory from source to destination.

To solve the formulation, we establish the classical A* algorithm as the foundational planner. Unlike blind search methods, A* directs the exploration by minimizing a heuristic-guided cost function $f(s)$ for every visited node $s$:
\begin{equation}
    f(s) = g(s) + h_{A^*}(s)
    \label{eq:astar_cost}
\end{equation}
Here, $g(s)$ represents the accumulated exact cost from $s_0$ to current node $s$, while $h_{A^*}(s)$ is a heuristic function estimating the remaining cost from $s$ to $s_g$.

The algorithm maintains a priority queue ($\mathcal{O}_\text{open}$) containing candidate nodes sorted by their $f$-values. At each iteration, the planner extracts the node with the minimal $f(s)$, expands its valid neighbors in $\mathcal{S}_{\text{free}}$, and relaxes their costs if a more efficient path is discovered. Expanded nodes are retired to a CLOSED set ($\mathcal{C}_\text{close}$) to prevent redundant processing. The search terminates when $s_g$ is selected for expansion, at which point the optimal trajectory is reconstructed by backtracking parent pointers. 
Crucially, the optimality of the solution is guaranteed provided that $h_{A^*}(s)$ is \textit{admissible} (i.e., it never overestimates the true remaining distance).

\subsection{Overview}
Figure~\ref{fig:schema} and Algorithm~\ref{alg:mmp-a-star} illustrates the overall workflow of the proposed \textsc{MMP-A*}, which improves upon LLM-A* to resolve its scalability constraints and enhance adaptability in complex scenarios. The planner operates in three sequential stages that couple linguistic reasoning with visual spatial grounding. 
First, the LLM analyzes the global map and proposes a set of coarse waypoints representing high-level intent and directional guidance (Section \ref{sec:LLM-Guided Waypoint Generation}). 
Second, the VLM refines these suggestions by examining the visual map, removing waypoints that lie in blocked regions or near walls, thus ensuring geometric feasibility and free-space alignment (Section \ref{sec:VLM-Based Visual Refinement}). 
Finally, the refined waypoints are injected into the A* search as a multimodal prior within an adaptive fading-checkpoint heuristic (Section \ref{sec:Adaptive Heuristic Search Integration}). 
This integration enables the search to leverage early linguistic cues for fast exploration while progressively decaying their influence to guarantee admissibility and optimal convergence.

\subsection{LLM-Guided Waypoint Generation}
\label{sec:LLM-Guided Waypoint Generation}
In this phase, the LLM processes a textual encoding of the environment map to propose a high-level navigational strategy. The model is tasked with generating a \textit{target list} $T$, comprising coarse-grained waypoints that bridge the start state $s_0$ and the goal state $s_g$, thereby leveraging the LLM's global reasoning capabilities. To ensure the structural integrity of the proposed path, two fundamental constraints are enforced:

\begin{enumerate}
    \item \textbf{Endpoint Consistency:} 
    The target set must strictly delimit the trajectory. Formally, we require $\{s_0, s_g\} \subseteq T$. If the model fails to generate these boundary states, they are explicitly appended to the set.

    \item \textbf{Feasibility Verification:} 
    Every generated waypoint $t \in T$ is validated against the environmental constraints. Any waypoint localized within an obstacle region (i.e., $t \in \mathcal{S}_{obs}$) is identified as invalid and pruned from the list prior to subsequent processing.
\end{enumerate}

Despite these structural safeguards, the LLM operates on abstract textual data and lacks intrinsic spatial grounding. Consequently, the generated waypoints may exhibit geometric inaccuracies. Critically, as the environment scales or obstacle density increases, the textual representation becomes unwieldy, leading to prohibitive computational costs and context-length bottlenecks. These limitations necessitate the subsequent stage, which employs visual perception to robustly validate and optimize the waypoint candidates.

\subsection{VLM-Based Visual Refinement}
\label{sec:VLM-Based Visual Refinement}
To mitigate spatial hallucinations from the text-based phase, we introduce a VLM refinement stage that validates the coarse list $T$. The model is prompted with two aligned visual inputs: (i) a \textit{raw occupancy grid} marking barriers and $\{s_0, s_g\}$, and (ii) a \textit{visualization layer} overlaying LLM-proposed waypoints onto the grid.
Our prompt engineering operationalizes the concept of a valid checkpoint as a node that maintains safety margins from walls and avoids congestion. The refinement workflow proceeds as follows:

\begin{enumerate}
    \item \textbf{Global Scene Understanding:} The VLM scans the raw grid to map the global structure of the maze, assessing the spatial relationship between obstacles and the potential path from $s_0$ to $s_g$.

    \item \textbf{Feasibility Filtering:} The model scrutinizes each candidate in $T$ against the visual evidence. Waypoints located in open, strategic positions are preserved. In contrast, those identified in blocked regions, dead-ends, or geometrically constrained passages are flagged as hazardous or redundant and are subsequently discarded to ensure a robust search space.
\end{enumerate}

We prioritize visual verification over direct generation to exploit the resolution independence of the top-down view. Unlike coordinate-based generation, which struggles with the combinatorial explosion of large grids, a visual approach allows us to abstract the map into a fixed-size image. This ensures that the VLM's ability to interpret global topology remains robust regardless of the map's actual scale. 


\subsection{Adaptive Heuristic Search Integration}
\label{sec:Adaptive Heuristic Search Integration}
The original LLM-A* approach relies on waypoints generated by an LLM to guide the A* search, which significantly enhances computational efficiency. However, this mechanism introduces a strong dependency on the correctness of the waypoints: when a waypoint is misplaced or misleading, the heuristic estimation
\begin{equation}
    h_{LLM-\!A^*}(n) = h_{A^*}(n) + cost(n, t_k)
\end{equation}
becomes unstable. Here $cost(n, t_k)$ measures the cost from the current node $n$ to the current waypoint $t_k$. Consequently, the search tends to diverge toward suboptimal areas, greatly increasing computation and slowing convergence. This limitation arises from the static treatment of LLM-generated waypoints, where the algorithm maintains excessive confidence in each waypoint regardless of its reliability.
To address this issue, we introduce an \textit{adaptive decay factor} $\alpha$ to dynamically reduce the influence of LLM-generated waypoints over time. Specifically, the heuristic function is reformulated as
\begin{equation}
    h_{MMP-\!A^*}(n) = h_{A^*}(n) + \alpha^{k} cost(n, t_k),
\end{equation}
where $\alpha \in (0, 1)$ decays exponentially with each waypoint switch ($k$ denotes the index of the current waypoint).

As the search progresses toward the goal, the influence of intermediate waypoints gradually diminishes, making the heuristic increasingly focused on the actual target. This adaptive strategy mitigates the bias introduced by unreliable LLM waypoints, improving both the stability of heuristic estimation and the overall efficiency of the search process.


\subsection{Theoretical Analysis}
\label{subsec:theoretical}

We analyze the theoretical properties of the adaptive decay heuristic employed in MMP-A*.

\begin{proposition}[Bounded Suboptimality]
Let $h_{A^*}(n)$ be an admissible heuristic and define
\[
h_{\text{MMP-A}^*}(n) = h_{A^*}(n) + \alpha^k \cdot c(n,t_k),
\quad \alpha \in (0,1),
\]
where $t_k$ denotes the current waypoint and $c(n,t_k)$ is the estimated cost from node $n$ to $t_k$. Assume that there exists a finite constant $D_{\max}$ such that $c(n,t_k) \le D_{\max}$ for all $n$ and $k$, e.g., the maximum shortest-path distance in the finite grid. Then, for any finite $k$,
\[
h_{\text{MMP-A}^*}(n) \le h^*(n) + \alpha^k D_{\max},
\]
where $h^*(n)$ denotes the true optimal cost-to-go. Consequently, the heuristic exhibits bounded additive suboptimality with an error term $\alpha^k D_{\max}$.
\end{proposition}

\paragraph{Proof Sketch.}
By admissibility of $h_{A^*}$, we have $h_{A^*}(n) \le h^*(n)$. Since the waypoint-dependent term $c(n,t_k)$ is bounded by $D_{\max}$ and scaled by $\alpha^k$, the heuristic introduces a bounded additive overestimation that decays exponentially with the waypoint index. \hfill $\square$

\begin{proposition}[Pointwise Convergence]
For any node $n$,
\[
\lim_{k \to \infty} h_{\text{MMP-A}^*}(n) = h_{A^*}(n).
\]
\end{proposition}

\paragraph{Proof Sketch.}
Since $\alpha \in (0,1)$, the decay factor satisfies $\lim_{k \to \infty} \alpha^k = 0$. As a result, the waypoint guidance term vanishes asymptotically, and the heuristic converges pointwise to the admissible base heuristic $h_{A^*}$. \hfill $\square$

\begin{proposition}[Robustness against Long-Horizon Spatial Hallucinations]
In large-scale environments where the sequence length $k$ is significant, VLM reliability typically degrades, leading to high-error waypoints (hallucinations).
Under the LLM-A* formulation ($\alpha = 1$), a misleading waypoint $t_k$ imposes a persistent heuristic penalty $c(n, t_k)$, forcing the planner to detour towards the erroneous location regardless of the path length.
In contrast, MMP-A* attenuates this risk via the decay factor $\alpha^k$. For large $k$, the influence of the misleading waypoint vanishes:
\[
\lim_{k \to \infty} \alpha^k \cdot c(n, t_k) = 0.
\]
\end{proposition}
\noindent This property ensures that as the problem scale increases, MMP-A* automatically decouples from potentially unreliable VLM guidance and reverts to the admissible goal-directed behavior of $h_{A^*}(n)$, effectively bypassing misleading traps that would otherwise entrap a LLM-A* planner.


%% file: sections/4_experiments.tex
\section{Experiments}
\begin{table*}[htbp]
\centering
\begin{adjustbox}{max width=\textwidth}
\footnotesize
\begin{tabular}{l l l c c c c}
\toprule
\textbf{Methodology} & \textbf{LLM Model} & \textbf{VLM Model} &
\textbf{Operation Ratio ↓ (\%)} & \textbf{Storage Ratio ↓ (\%)} &
\textbf{Relative Path Length ↓ (\%)} & \textbf{Valid Path Ratio ↑ (\%)} \\
\midrule
\textbf{A*} &  &  & 100 & 100 & 100 & 100 \\
\midrule
\multirow{4}{*}{\textbf{LLM-A*}}
 & DeepSeek-V3     &  & 125.2 & 114.6 & 103.4 & 100 \\
 & Llama-3.3-70B   &  & 112.1 & 103.1 & 103.3 & 100 \\
 & Qwen2.5-7B      &  & 191.6 & 141.4 & 102.5 & 100 \\
 & GPT-4o-mini     &  & 115.3 & 104.6 & 103.5 & 100 \\
\midrule
\multirow{12}{*}{\textbf{MMP-A*}}
 &  & Gemma-3n-E4B    & 91.1 & 86.0 & 101.9 & 100 \\
 & DeepSeek-V3     & Llama 4 Maverick & 93.9 & 87.8 & 101.6 & 100 \\
 &  & Qwen2.5-VL       & 84.9 & 80.6 & 102.2 & 100 \\
\cmidrule(lr){2-7}
 &  & Gemma-3n-E4B    & 88.5 & 81.5 & 102.0 & 100 \\
 & Llama-3.3-70B   & Llama 4 Maverick & 97.0 & 91.0 & 101.8 & 100 \\
 &  & Qwen2.5-VL       & \textbf{81.0} & \textbf{76.0} & 102.3 & 100 \\
\cmidrule(lr){2-7}
 &  & Gemma-3n-E4B    & 150.4 & 106.3 & 101.9 & 100 \\
 & Qwen2.5-7B      & Llama 4 Maverick & 181.0 & 129.8 & \textbf{101.5} & 100 \\
 &  & Qwen2.5-VL       & 162.8 & 114.8 & 102.2 & 100 \\
\cmidrule(lr){2-7}
 &  & Gemma-3n-E4B    & 82.4 & 76.5 & 102.2 & 100 \\
 & GPT-4o-mini     & Llama 4 Maverick & 97.4 & 88.3 & 101.6 & 100 \\
 &  & Qwen2.5-VL       & \textbf{81.0} & 76.4 & 102.3 & 100 \\
\bottomrule
\end{tabular}
\end{adjustbox}
\caption{\small Quantitative comparison between baseline A*, LLM-A* and our multimodal framework MMP-A*}
\label{tab:experiment}
\vspace{-0.4cm}
\end{table*}
\subsection{Dataset} \label{sec:dataset}
\begin{figure}[!tb]
	\centering
	\includegraphics[width=1\linewidth] {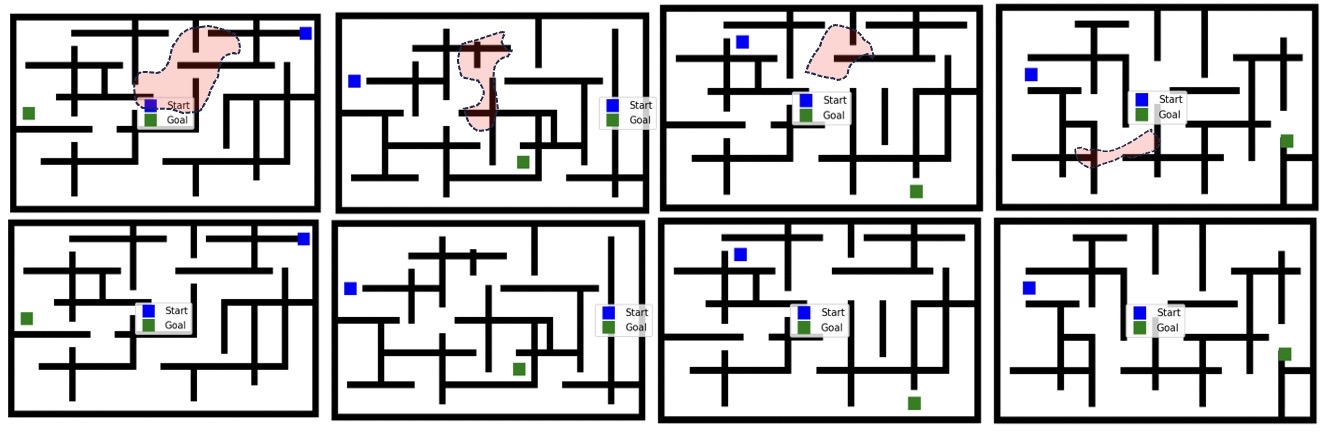}
\caption{\small{\textbf{Visualization of the experimental setup.} The bottom row depicts the core dataset featuring dense, maze-like topologies. The top row illustrates the generality assessment subset, where irregular, amorphous barriers (shaded red) are superimposed to rigorously evaluate the framework's visual generalization capabilities against non-geometric obstacles.}}
    \label{high-complexity path planning dataset}
\end{figure}

We validate our framework using a suite of 200 high-complexity maps (100×60) that exhibits significantly greater topological intricacy than the sparse environments typical of prior \textsc{LLM-A*} research. By incorporating labyrinthine corridors and deceptive dead-ends, these topologies are explicitly designed to induce local optima, effectively neutralizing greedy heuristics and necessitating robust global reasoning.
To ensure a holistic assessment across diverse navigational scenarios, we implement a dual-faceted evaluation protocol:
\begin{enumerate}
    \item \textbf{Scalability} is examined by expanding map dimensions from 30×50 to 240×400 under strict topological invariance, explicitly isolating algorithmic efficiency from environmental structure.
    \item \textbf{Environmental Complexity} is modulated via obstacle density, culminating in Level 5, a rigorous benchmark featuring approximate 12 intertwined barrier clusters and deep dead-ends designed to challenge multimodal reasoning.
    \item \textbf{Irregular Obstacles} introduces undefined hazard zones that do not fit standard grid lines. This evaluates visual perception in scenarios where text-based descriptions become inefficient or impractical.
\end{enumerate}
Figure \ref{high-complexity path planning dataset} illustrates samples from our curated datasets.
For a detailed specification of these parameters and generation protocols, please refer to Appendix~\ref{Details of Dataset Construction}.

\subsection{Experimental Setup}
\paragraph{Models and Parameters.}
We evaluate \textsc{MMP-A*} using representative models including GPT-4o-mini, Llama-3.3-70B, Qwen2.5-7B, and DeepSeek-V3 for reasoning, alongside Llama-4-Maverick, Gemma-3n, and Qwen2.5-VL for perception. We focus our evaluation on A* and \textsc{LLM-A*} as primary baselines to isolate and validate the specific efficiency gains of the VLM-integrated methodology. Implementation-wise, we adopt standard few-shot, Chain-of-Thought, and Recursive Path Evaluation (RePE) strategies with a fixed heuristic decay factor \(\alpha = 0.7\) (see Appendix~\ref{Prompts in LLMs} and \ref{Prompt in VLM} for full prompt details).

\paragraph{Experiment Metrics and Objectives.}
Our evaluation focuses on efficiency and scalability by benchmarking \textsc{MMP-A*} against A* and LLM-A*. We assess computational cost and memory usage via Operation and Storage Ratios, computed as the geometric mean of performance ratios between \textsc{MMP-A*} and the A* baseline (\(\frac{\text{MMP-A*}}{\text{A*}}\)), where lower values indicate superior resource efficiency. Path optimality is evaluated via Relative Path Length, while system reliability is measured by the Valid Path Ratio, representing the proportion of successfully generated collision-free trajectories. Comprehensive results are summarized in Table~\ref{tab:experiment}. 
Figure \ref{Visual comparison} in Appendix visualizes representative sample results, demonstrating how MMP-A* allows the planner to bypass misleading waypoints and yield smoother trajectories.

\subsection{Experimental Results}\label{sec:result}
\paragraph{Overall Quantitative Results.}

\begin{table}[!h]
\centering
\footnotesize
\resizebox{\columnwidth}{!}{%
\begin{tabular}{l l l c c}
\toprule
\textbf{Methodology} & \textbf{LLM Model} & \textbf{VLM Model} & \textbf{Rel. Path $\downarrow$} & \textbf{Valid $\uparrow$} \\
\midrule
\textbf{A*} &  &  & 100 & 100 \\
\midrule
\multirow{4}{*}{\textbf{LLM}} 
 & DeepSeek-V3   &  & 116.8 & 7.0 \\
 & Llama-3.3-70B &  & 123.5 & 8.6 \\
 & Qwen2.5-7B    &  & 130.4 & 4.5 \\
 & GPT-4o-mini   &  & 112.0 & 7.5 \\
\midrule
\multirow{3}{*}{\textbf{VLM}} 
 &  & Gemma-3n-E4B   & 114.3 & 6.0 \\
 &  & Llama 4 Maverick & 122.4 & 6.5 \\
 &  & Qwen2.5-VL       & 119.5 & 8.0 \\
\midrule
\multirow{12}{*}{\textbf{LLM + VLM}} 
 & \multirow{3}{*}{DeepSeek-V3} & Gemma-3n-E4B   & 110.0 & 8.0 \\
 &                                & Llama 4 Maverick & 110.6 & 9.0 \\
 &                                & Qwen2.5-VL       & 107.9 & 9.5 \\
 \cmidrule(lr){2-5} 
 & \multirow{3}{*}{Llama-3.3-70B} & Gemma-3n-E4B   & 101.3 & 5.1 \\
 &                                & Llama 4 Maverick & 114.9 & 9.7 \\
 &                                & Qwen2.5-VL       & 105.7 & 6.1 \\
 \cmidrule(lr){2-5}
 & \multirow{3}{*}{Qwen2.5-7B}    & Gemma-3n-E4B   & 112.6 & 4.0 \\
 &                                & Llama 4 Maverick & 118.0 & 5.0 \\
 &                                & Qwen2.5-VL       & 109.1 & 3.5 \\
 \cmidrule(lr){2-5}
 & \multirow{3}{*}{GPT-4o-mini}   & Gemma-3n-E4B   & 100.9 & 5.5 \\
 &                                & Llama 4 Maverick & 104.2 & 6.0 \\
 &                                & Qwen2.5-VL       & 103.3 & 6.0 \\
\bottomrule
\end{tabular}
}
\caption{\small Comparison between A* and GenAI-based methods}
\label{tab:genai}
\end{table}
The comprehensive evaluation in Table~\ref{tab:experiment}--\ref{tab:genai} reveals that purely generative models fail to capture topological constraints of complex environments, yielding Valid Path Ratios below $10\%$ and trajectories inflated by up to $30\%$ due to geometric hallucinations.
While integrating language guidance with search via \textsc{LLM-A*} resolves the validity issue, it inadvertently incurs a substantial computational penalty, with operation ratios surging to $191.6\%$ in some configurations as the search algorithm is forced to explore misleading regions suggested by noisy heuristics. In stark contrast, our proposed \textsc{MMP-A*} achieves a superior balance of efficiency and optimality across all model backbones. By leveraging VLM-based visual pruning to eliminate infeasible waypoints before the search commences and employing an adaptive heuristic decay, \textsc{MMP-A*} not only guarantees $100\%$ validity but also drastically reduces resource consumption, bringing operation and storage costs down to $81.0\%$ and $76.0\%$ respectively while maintaining path lengths within $2\%$ of the optimal solution.
We discuss the comparative efficiency of LLM–VLM models and their limitations in Appendix ~\ref{limitation}.
\begin{table}[!h]
\centering
\small
\begin{tabular}{l c c c} 
\toprule
\textbf{Methodology} & \textbf{Operation $\downarrow$} & \textbf{Storage $\downarrow$}
& \textbf{Rel. Path $\downarrow$}\\
\midrule
A* & 100 & 100 & 100 \\
LLM-A* & 123.7 & 113.1 & 103.2 \\ 
MMP-A* & \textbf{88.9} & \textbf{82.1} & \textbf{102.1} \\ 
\bottomrule
\end{tabular}
\caption{\small Comparison of A*, LLM-A*, and MMP-A* performance.}
\label{tab:summary_comparison}
\end{table}
\paragraph{Complex Environments Analysis} The robustness of this improvement becomes clearer in Table~\ref{tab:complex} and Figure~\ref{fig:increment}a, where we stress-test the system under increasingly complex maze environments. As obstacle density grows, \textsc{LLM-A*} exhibits unstable behavior, often consuming two to three times more operations than the baseline due to misleading guidance from noisy checkpoints. In contrast, \textsc{MMP-A*} maintains steady performance across all difficulty levels. For instance, with the DeepSeek-V3 and Qwen2.5-VL pairing, the operation ratio remains between 45--97\% and the storage ratio between 53--86\% from Level~1 to Level~5, consistently outperforming both A* and \textsc{LLM-A*}. The relative path length stays close to optimal, indicating that the efficiency gain arises from reduced exploration rather than from sacrificing solution quality. These results show that visual pruning is especially useful in cluttered or irregular environments, as it stops the search from following dead-end or wall-hugging paths suggested by the LLM.

\paragraph{Scalability Analysis.}
Table~\ref{tab:scale} and Figure~\ref{fig:increment}b further investigates scalability across grid sizes ranging from $30{\times}50$ to $240{\times}400$. As observed, \textsc{LLM-A*} becomes increasingly unstable at larger scales, sometimes performing worse than vanilla A* due to accumulated waypoint noise over longer planning horizons. Conversely, \textsc{MMP-A*} demonstrates consistent and even improving efficiency with scale. Using the GPT-4o-mini and Gemma-3n-E4B pairing, operation ratios rise only slightly from about 63\% to 77\%, while storage ratios decline from 73\% to 59\% as the map enlarges, all while preserving near-optimal path lengths. This stable scaling behavior confirms that our multimodal filtering and fading-heuristic design not only improves small-scale navigation but also generalizes to large, complex search spaces. 
\begin{table*}[htbp]
\centering
\begin{adjustbox}{max width=\textwidth}
\footnotesize
\begin{tabular}{l l *{5}{ccc}}
\toprule
\multirow{2}{*}{\textbf{Methodology}} & \multirow{2}{*}{\textbf{Base Model}} &
\multicolumn{3}{c}{\textbf{Level 1}} &
\multicolumn{3}{c}{\textbf{Level 2}} &
\multicolumn{3}{c}{\textbf{Level 3}} &
\multicolumn{3}{c}{\textbf{Level 4}} &
\multicolumn{3}{c}{\textbf{Level 5}} \\
\cmidrule(lr){3-5}\cmidrule(lr){6-8}\cmidrule(lr){9-11}\cmidrule(lr){12-14}\cmidrule(lr){15-17}
 &  & \textbf{Op} & \textbf{Stor} & \textbf{Rel} & \textbf{Op} & \textbf{Stor} & \textbf{Rel}
 & \textbf{Op} & \textbf{Stor} & \textbf{Rel} & \textbf{Op} & \textbf{Stor} & \textbf{Rel}
 & \textbf{Op} & \textbf{Stor} & \textbf{Rel} \\
\midrule
\textbf{A*} & -- 
& 100 & 100 & 100 & 100 & 100 & 100 & 100 & 100 & 100 & 100 & 100 & 100 & 100 & 100 & 100 \\
\midrule
\multirow{4}{*}{\textbf{LLM-A*}}
  & DeepSeek-V3              & 52.1 & 59.3 & 103.6 & 70.9 & 72.6 & 104.0 & 130.5 &  103.0 & 105.5 & 110.0 & 96.8 & 103.7 & 124.1 & 109.6 & 103.1 \\
  & Llama-3.3-70B   & 86.6 & 62.5 & 103.5 & 68.7 & 72.2 & 103.3 & 119.8 & 88.7 & 104.8 & 99.4 & 94.3 & 104.7 & 112.2 & 103.7 & 102.7 \\
  & Qwen2.5-7B      & 300.0 & 128.1 & 103.8 & 130.6 & 116.5 & 102.6 & 300.2 & 139.1 & 104.1 & 149.7 & 125.2 & 103.7 & 121.6 & 112.0 & 102.8 \\
  & GPT-4o-mini              & 74.6 & 75.8 & 103.3 & 88.9 & 83.1 & 102.8 & 129.8 & 82.7 & 105.6 & 99.3 & 93.6 & 104.3 & 111.5 & 104.7 & 102.8 \\
\midrule
\multirow{4}{*}{\textbf{MMP-A*}}
  & DeepSeek-V3 + Qwen2.5-VL   & \textbf{45.4} & \textbf{53.1} & 102.3 & 61.7 & 65.2 & 103.0 & \textbf{81.8} & 78.4 & \textbf{102.9} & 78.2 & 73.9 & 102.9 & 96.8 & 86.2 & 102.1 \\
  & Llama-3.3-70B + Qwen2.5-VL & 65.9 & 56.8 & 102.5 & 66.4 & 68.8 & 102.6 & 83.3 & \textbf{70.3} & 104.0 & 75.5 & 75.8 & 103.3 & 90.5 & 80.8 & 102.1 \\
  & Qwen2.5-7B + Qwen2.5-VL    & 269.8 & 102.7 & 101.9 & 80.9 & 78.7 & 102.1 & 221.2 & 114.2 & 103.2 & 111.2 & 98.4 & 102.8 & 111.8 & 99.1 & 102.4 \\
  & GPT-4o-mini + Qwen2.5-VL           & 49.5 & 60.0 & \textbf{101.7} & \textbf{44.3} & \textbf{52.3} & \textbf{101.9} & 93.9 & 74.3 & 103.0 & \textbf{71.6} & \textbf{71.0} & \textbf{102.2} & \textbf{88.9} & \textbf{84.7} & \textbf{101.9} \\
\bottomrule
\end{tabular}
\end{adjustbox}
    \caption{Complex Environment Experiment}
    \label{tab:complex}
    \vspace{-0.4cm}
\end{table*}

\begin{table*}[htbp]
\centering
\begin{adjustbox}{max width=\textwidth}
\footnotesize
\begin{tabular}{l l *{5}{ccc}}
\toprule
\multirow{2}{*}{\textbf{Methodology}} & \multirow{2}{*}{\textbf{Base Model}} &
\multicolumn{3}{c}{\textbf{Level 1} (30$\times$50)} &
\multicolumn{3}{c}{\textbf{Level 2} (60$\times$100)} &
\multicolumn{3}{c}{\textbf{Level 3} (120$\times$200)} &
\multicolumn{3}{c}{\textbf{Level 4} (180$\times$300)} &
\multicolumn{3}{c}{\textbf{Level 5} (240$\times$400)} \\
\cmidrule(lr){3-5}\cmidrule(lr){6-8}\cmidrule(lr){9-11}\cmidrule(lr){12-14}\cmidrule(lr){15-17}
 &  & \textbf{Op} & \textbf{Stor} & \textbf{Rel} & \textbf{Op} & \textbf{Stor} & \textbf{Rel}
 & \textbf{Op} & \textbf{Stor} & \textbf{Rel} & \textbf{Op} & \textbf{Stor} & \textbf{Rel}
 & \textbf{Op} & \textbf{Stor} & \textbf{Rel} \\
\midrule
\textbf{A*} & -- 
& 100 & 100 & 100 & 100 & 100 & 100 & 100 & 100 & 100 & 100 & 100 & 100 & 100 & 100 & 100 \\
\midrule
\textbf{LLM-A*} & GPT-4o-mini
& 85.5 & 88.5 & 101.5 & 97.2 & 91.3 & 103.4 & 102.8 & 92.9 & 104.8 & 106.1 & 93.7 & 103.7 & 100.7 & 83.1 & 104.1 \\
\midrule
\textbf{MMP-A*} & GPT-4o-mini + Gemma-3n-E4B
& \textbf{63.1} & \textbf{73.4} & \textbf{100.9} & \textbf{80.5} & \textbf{76.9} & \textbf{101.8} & \textbf{89.9} & \textbf{79.8} & \textbf{102.2} & \textbf{89.4} & \textbf{75.4} & \textbf{102.6} & \textbf{76.6} & \textbf{59.3} & \textbf{102.5} \\
\bottomrule
\end{tabular}
\end{adjustbox}
    \caption{Scale Robustness Experiment}
    \label{tab:scale}
    \vspace{-0.4cm}
\end{table*}
\paragraph{General Applicability beyond Grid Constraints.}
While specialized A* variants offer runtime acceleration, they remain confined to rigid grid structures, and \textsc{LLM-A*} struggles to textually describe amorphous hazards (e.g., irregular hazardous regions), \textsc{MMP-A*} achieves robust generalization through direct visual perception. By treating the environment as a visual scene rather than a coordinate list, our VLM-based approach intuitively interprets arbitrary obstacle shapes that defy rigid geometric definitions. This versatility leads to superior efficiency (Table~\ref{tab:summary_comparison}), with \textsc{MMP-A*} achieving an operation score of 88.9, effectively navigating complex scenarios where traditional grid-centric assumptions break down.

\subsection{Ablation Study}\label{sec:ablation}
\paragraph{Alpha-Decay Sensitivity Analysis.}
As shown in Table~\ref{tab:adaptive} in Appendix and Figure~\ref{fig:alpha-decay}, incorporating adaptive decay slightly increases the operation and storage ratios, by about 5–10\%, since the planner performs extra expansions to refine its heuristic. Nevertheless, it remains markedly more efficient than vanilla A*, while consistently producing smoother and shorter paths. This small overhead is offset by a 1–2\% reduction in relative path length, confirming that adaptive decay effectively stabilizes the search by gradually lowering reliance on uncertain early waypoints. When varying the decay coefficient $\alpha$, we observe a clear efficiency–optimality trade-off: higher $\alpha$ values reduce operation cost but yield longer paths due to stronger waypoint bias, whereas lower $\alpha$ values encourage broader exploration and shorter trajectories at slightly higher cost. Overall, moderate $\alpha$ values achieve the best balance, preserving efficiency gains while maintaining near-optimal path quality. 

\paragraph{Runtime–performance trade-off.} We analyze the runtime cost, inclusive of API latency, across varying environmental complexities. \textsc{MMP-A*} effectively amortizes reasoning overheads in large-scale settings, achieving runtime parity or even acceleration compared to \textsc{LLM-A*} while delivering superior memory efficiency. For a detailed discussion, please refer to Appendix~\ref{sec:runtime-trafeoff}.
\begin{figure}[!tb]
	\centering
	\includegraphics[width=1\linewidth] {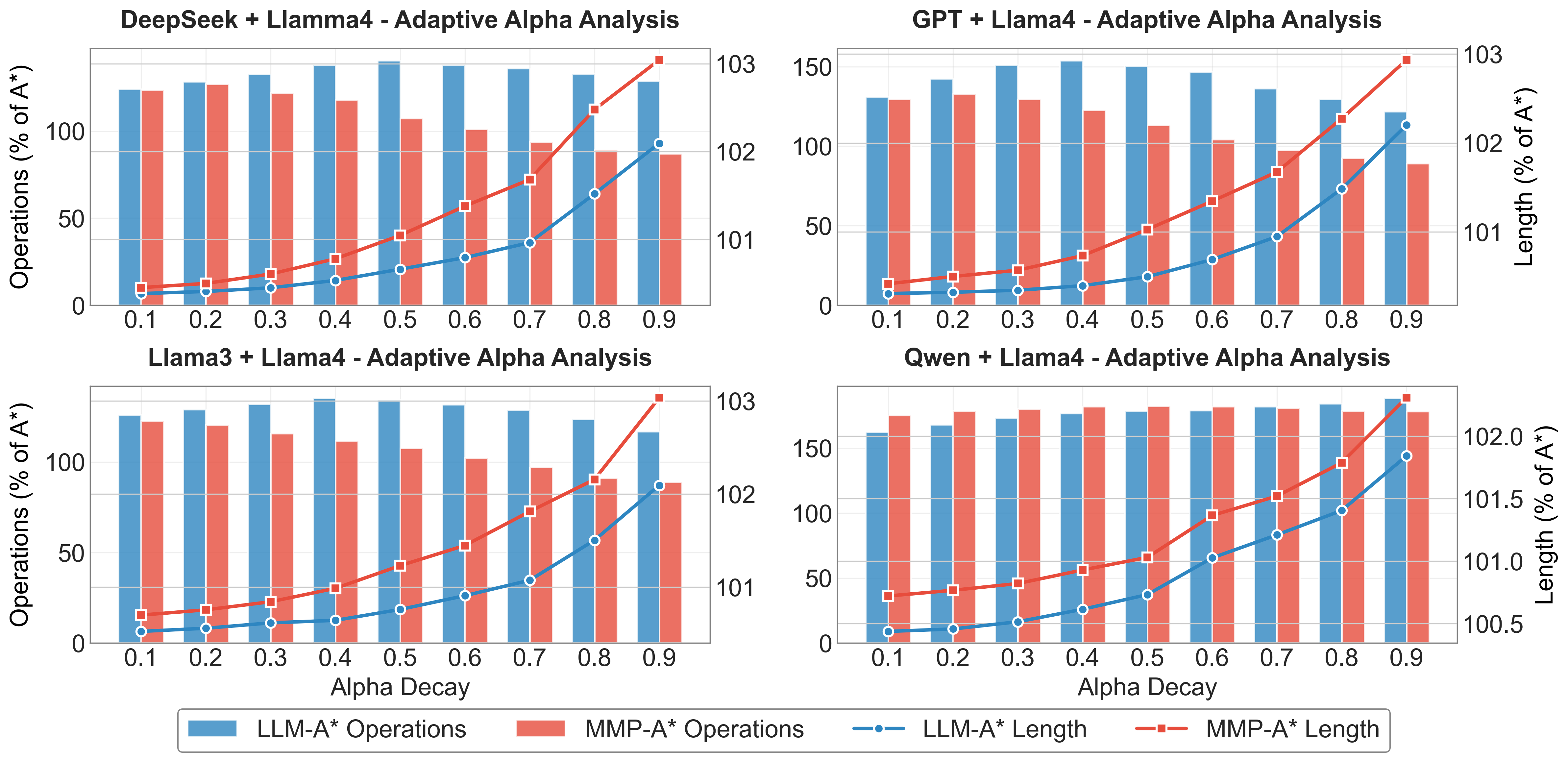}
\caption{
\textbf{Alpha-Decay Sensitivity Analysis.}
Operation ratio (bars) and relative path length (lines) of \textsc{LLM-A*} and \textsc{MMP-A*} under varying decay coefficients $\alpha$.
}
		\label{fig:alpha-decay}
\end{figure}

\paragraph{Prompt Engineering Strategy.}
We evaluated three prompting schemes to analyze their influence on waypoint generation and search efficiency. As summarized in Table~\ref{tab:prompt} and Figure \ref{prompt_compare} in Appendix, CoT prompts generally yield better reasoning consistency but incur higher computational cost, with operation ratios often 10–20\% above Few-Shot setups. RePE introduces recursive self-evaluation, which improves local waypoint precision and slightly reduces path length (about 0.5–1.0\%), though at the expense of additional memory usage. Among these, Few-Shot remains the most lightweight and stable baseline, while RePE offers the best trade-off between cost and accuracy when integrated with \textsc{MMP-A*}. 
\input{figures/increment}

%% file: figures/increment.tex

\begin{figure}[!tb]
	\centering
	\includegraphics[width=0.9\linewidth] {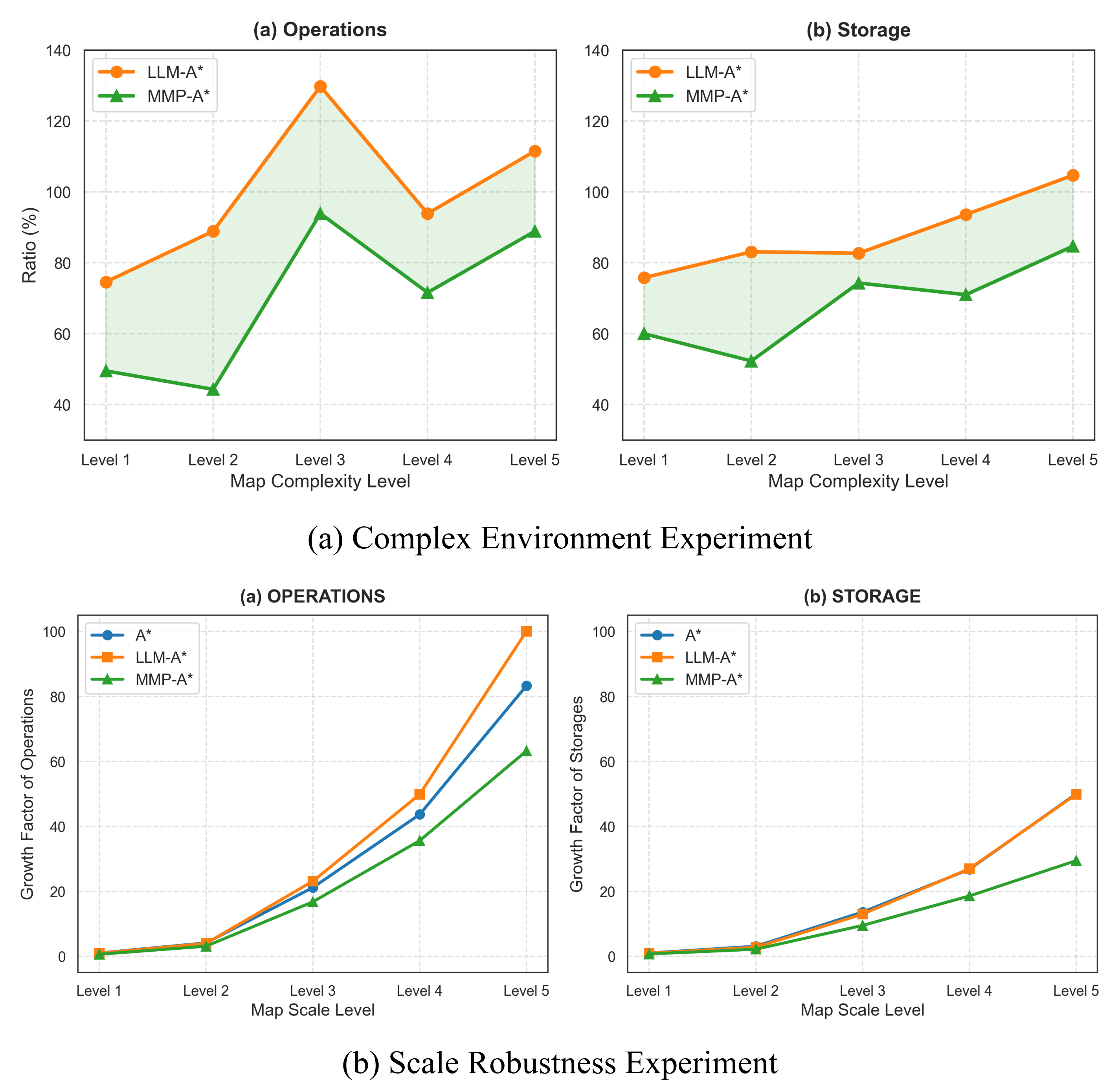}
\caption{
\small
\textbf{(a)~Complex Environment Experiment:} Operation and storage ratios across increasing map complexity levels.
\textbf{(b)~Scale Robustness Experiment:} Growth factors of operations and storages across varying map sizes.
} 
		\label{fig:increment}
\end{figure}

%% file: sections/5_conclusion.tex
\section{Conclusion}
In this work, we proposed \textbf{MMP-A*}, a multimodal perception–enhanced pathfinding framework that tightly integrates LLM reasoning, VLM visual filtering, and adaptive heuristic decay into classical A* search. The method demonstrates particular strength in complex and cluttered environments, where pure language-guided or traditional heuristic planners often fail. Through extensive experiments, MMP-A* consistently maintains 100\% path validity while achieving lower operation and storage costs, especially under high obstacle density and large-scale maps. The adaptive decay mechanism further stabilizes performance by balancing exploration and guidance, allowing the planner to refine its trajectory as search progresses. These results suggest that MMP-A* is especially well-suited for dynamic or visually ambiguous navigation tasks, where multimodal understanding and adaptive reasoning are essential for robust and efficient planning.
\cleardoublepage

%% file: sections/6_appendix.tex
\begin{figure*}[!tb]
	\centering
	\includegraphics[width=1\linewidth] {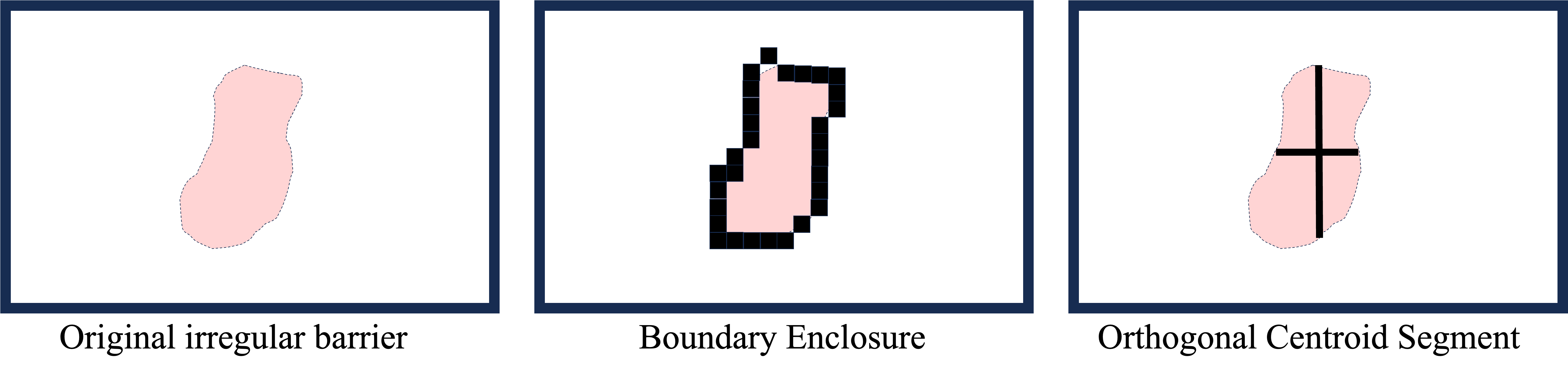}
\caption{
\textbf{Comparison of symbolic representation strategies for irregular obstacles.} From left to right: raw geometric data of an impassable barrier $B \in \mathcal{S}_{obs}$; our proposed skeletonized representation using axis-aligned segments for a constant token footprint and efficient LLM reasoning; and conventional boundary-based encoding which suffers from token explosion and high spatial noise.
}
\label{obstacle_reprenst}
\end{figure*}
\input{tables/prompt}

\section{Selection of Symbolic Obstacle Representations}
\label{represntation barier}
The efficacy of LLM-based path planning is heavily dependent on how spatial constraints are encoded into text. We evaluate the trade-offs between two primary symbolic strategies: \textbf{Boundary Enclosure} and our proposed \textbf{Orthogonal Centroid Segments}, as can be seen in Figure \ref{obstacle_reprenst}. Traditional boundary enclosure, which involves listing the coordinates of all boundary cells or multiple bounding boxes to encapsulate an irregular shape, proves highly ineffective for LLMs. As the map resolution increases or obstacle geometry becomes more complex, this method leads to a token explosion, often exceeding the model's context window and increasing inference latency. Furthermore, the high density of numerical coordinates creates significant spatial noise, causing the LLM to suffer from reasoning fatigue and hallucinations, where it fails to identify navigable gaps or misinterprets the overall topology of the environment.

In contrast, our MMP-A* framework adopts a skeletonized representation by approximating each irregular barrier $B$ with two axis-aligned orthogonal segments of unit width intersecting at the centroid $(x_c, y_c)$. This approach ensures a constant token footprint regardless of the obstacle's actual area or geometric complexity, allowing the system to scale efficiently to large-scale maps. By reducing a complex ``blob'' to two fundamental linear constraints, we enable the LLM to perform simplified logical checks (e.g., $x > x_c$ or $y < y_c$) more reliably. This provides the LLM with clear "topological hints" to understand which regions are impassable without overwhelming its processing capacity.

While this simplified representation does not capture every geometric nuance of an irregular obstacle, it is sufficient for the LLM's role as a high-level strategic planner. The inherent loss of detail is strategically compensated for in the subsequent VLM refinement stage. Unlike the LLM, the VLM operates on the original, un-abstracted map $\mathcal{M}$, where it perceives the precise boundaries of all barriers. The VLM refines the high-level waypoints generated by the LLM to ensure they are collision-free relative to the true obstacle geometry. This hierarchical division of labor allows the LLM to focus on global path topology while the VLM handles local geometric precision and safety assurance.

\begin{figure*}[!tb]
	\centering
	\includegraphics[width=1\linewidth] {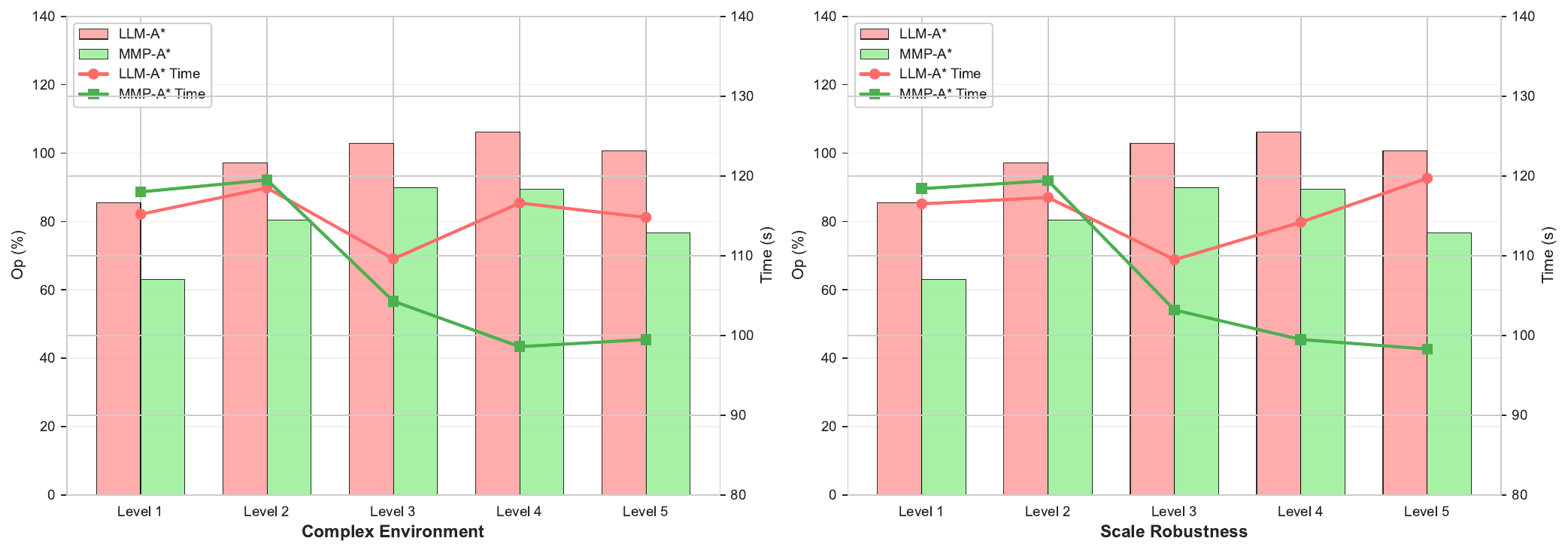}
\caption{
\textbf{Runtime Performance Trade-Off Between LLM-A* and MMP-A*.}
As environment complexity and map resolution increase, MMP-A* not only sustains superior optimality rates but also exhibits a decreasing runtime trend, 
becoming comparable to and even surpassing the runtime of classical A* at the highest complexity and resolution levels.
}
\label{runtime-trafeoff}
\end{figure*}

\section{Prompts in LLMs}
\label{Prompts in LLMs}
This appendix outlines the prompting techniques used in our MMP-A* algorithm to generate paths between start and goal points while navigating around obstacles. We employed different prompting strategies to evaluate their effectiveness in guiding the model. Below are the details of each technique along with the templates used.

\subsection{Standard 5-Shot Demonstration}

In the standard 5-shot demonstration in Table \ref{fig:fewshot}, the model is provided with five examples (or demonstrations) to guide the generation of the path. Each example includes start and goal points, along with horizontal and vertical barriers. The model is prompted to generate a path by following the pattern observed in the examples.

\subsection{Chain of Thought (CoT) Prompting}

The chain of thought prompting technique in Table \ref{fig:cot} provides a sequence of reasoning steps that the model follows to arrive at the final path. This technique includes a detailed thought process and evaluation for each step, helping the model to understand the rationale behind the path generation.

\subsection{Recursive Path Evaluation (RePE)}

In the recursive path evaluation technique shown Table \ref{fig:repe}, the model iteratively evaluates the path at each step and makes decisions based on previous iterations. This process involves selecting points, evaluating their effectiveness, and adjusting the path as necessary to avoid obstacles and reach the goal.

\section{Prompt in VLM}
\label{Prompt in VLM}

This appendix outlines the visual–language prompting strategy employed in our \textsc{MMP-A*} framework to evaluate the spatial validity of LLM-generated waypoints. 
While the LLM module proposes candidate checkpoints in textual form, the VLM serves as a perception-based verifier that inspects corresponding maze images to filter out invalid, wall-touching, or redundant waypoints. 
This process ensures that only geometrically feasible and visually consistent landmarks remain before the A* search begins, thereby improving path reliability and reducing unnecessary node expansions.

Specifically, the VLM receives two paired images: the first is a clean map with start and goal points, and the second contains the same map annotated with candidate waypoints along a suggested blue route. 
The prompt, shown in Table~\ref{fig:vlm_template}, instructs the model to reason over the obstacle layout, assess visibility and clearance, and output a structured JSON object that lists only the selected waypoints to keep. 

\section{Details of Dataset Construction}
\label{Details of Dataset Construction}
The dataset for A* path planning is generated using a custom Python script, leveraging several key packages for randomization, geometric manipulation, visualization, and data management. The process involves the following steps:

\begin{enumerate}
    \item \textbf{Initialization}: The script initializes with specified map dimensions (x and y boundaries) and parameters (number of barriers and obstacles) for the number of unique environments and start-goal pairs.
    
    \item \textbf{Environment Creation}: For each map configuration, do the following:
    \begin{itemize}
        \item Random obstacles, horizontal barriers, and vertical barriers are generated within defined $x$ and $y$ ranges. For irregular barriers, we manually designed non-standard geometries by delineating their boundaries with dashed, semi-transparent red contours. A reference grid was then superimposed over these shapes to precisely determine their centroids $(x_c, y_c)$. This manual annotation process ensures that each irregular obstacle is accurately mapped to its simplified symbolic representation, providing a consistent ground truth for the LLM's axis-aligned orthogonal segment approximation.
        \item Start and goal points are randomly placed on the map, ensuring they do not intersect with any obstacles. Valid pairs form non-intersecting line segments.
    \end{itemize}
    
    \item \textbf{Data Storage}: The generated environments, including the obstacles and start-goal pairs, are stored in JSON format.
    
    \item \textbf{Query Generation}: Natural language queries are appended to each start-goal pair. These queries describe the task of finding a path that avoids the obstacles, which is supported as text input for LLMs.
    
    \item \textbf{Visualization}: The environments are visualized using \texttt{matplotlib}, displaying the grid, obstacles, and paths. The plots are supported to be saved as image files for reference and stream in a show..
\end{enumerate}

\begin{figure}[!tb]
	\centering
	\includegraphics[width=1\linewidth] {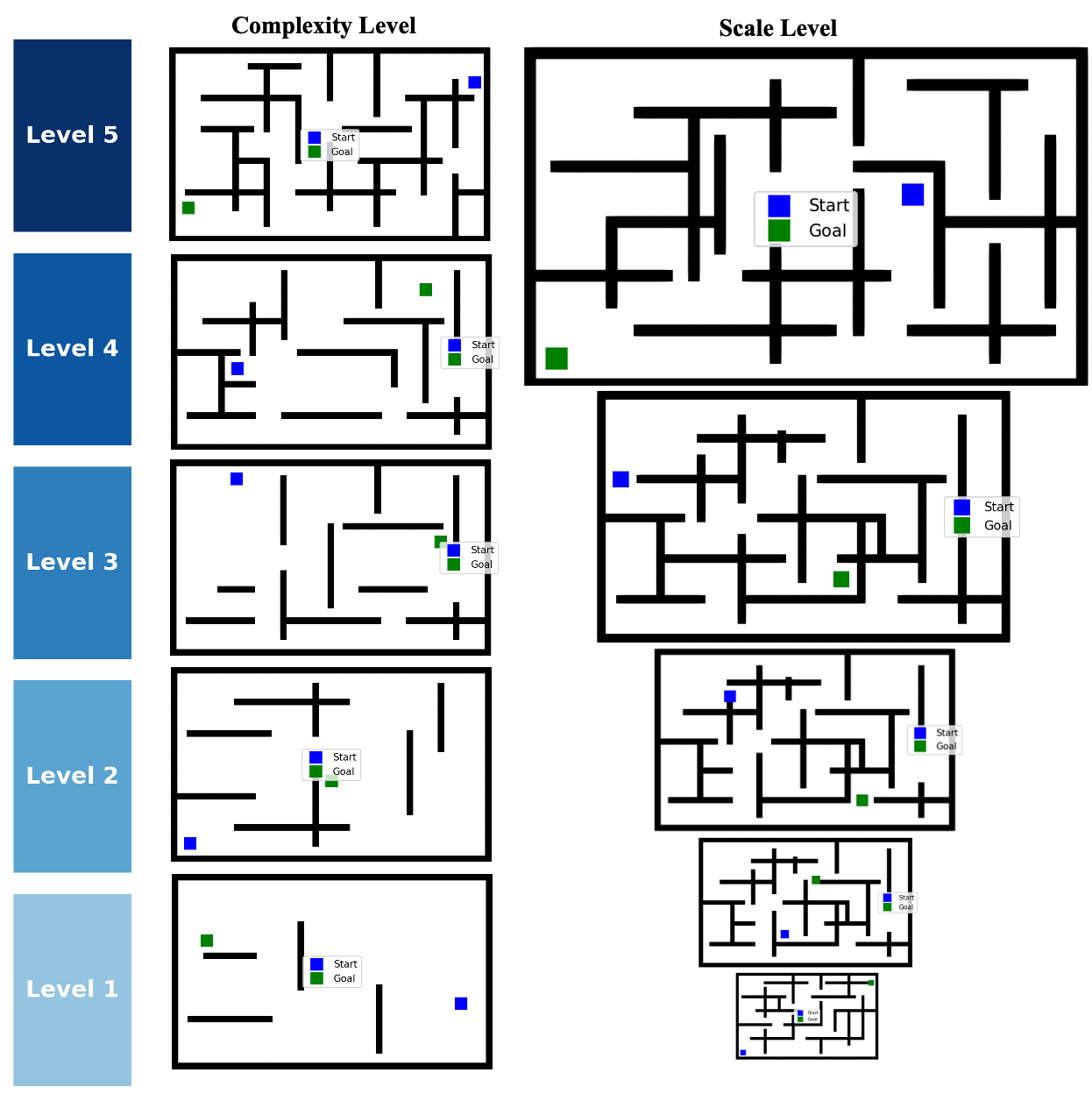}
\caption{
Hierarchical maze dataset for evaluating scalability and complexity in \textbf{MMP-A*}. 
Map sizes increase from $30\times50$ to $240\times400$, while barrier count rises from sparse layouts to $\sim$12 obstacles in Level~5. 
Each level preserves the relative spatial structure, enabling controlled analysis from simple to highly cluttered environments.
}
		\label{fig:Hierarchical}
\end{figure}

To systematically evaluate both scalability and environmental complexity, we constructed a hierarchical dataset divided into five levels of difficulty, as shown in Fig \ref{fig:Hierarchical}. Each level corresponds to a distinct grid resolution and barrier configuration, enabling progressive testing of \textsc{MMP-A*} under increasing spatial and structural challenges.

\textbf{Scalability setup.} The environment sizes range from small $30\times50$ maps in Level~1 up to large $240\times400$ maps in Level~5. For each level, we proportionally scale the map dimensions while preserving the relative topology of obstacles and corridors. This ensures that the navigational patterns remain consistent across scales, allowing controlled analysis of algorithmic efficiency and memory growth as problem size increases.

\textbf{Complexity setup.} The structural complexity is defined by the number and arrangement of barrier columns and rows. The benchmark Level~5 map contains approximately twelve primary obstacles (both horizontal and vertical), forming multiple intertwined corridors and dead-end traps. Lower levels progressively reduce the number of barriers by $2$–$3$ per level, producing simpler configurations while maintaining the same relative spatial layout. This hierarchical design creates a smooth transition from sparse to cluttered environments, enabling evaluation of robustness, operation ratio, and path validity across a controlled complexity spectrum.

Such construction allows the dataset to simultaneously stress-test both computational scalability and topological reasoning. In particular, Level~5 serves as the canonical benchmark environment, capturing the full navigational difficulty expected in real-world cluttered maps.

\section{Comparative Efficiency of LLM–VLM Models and Limitation}
\label{limitation}

\begin{figure*}[!tb]
	\centering
	\includegraphics[width=1\linewidth] {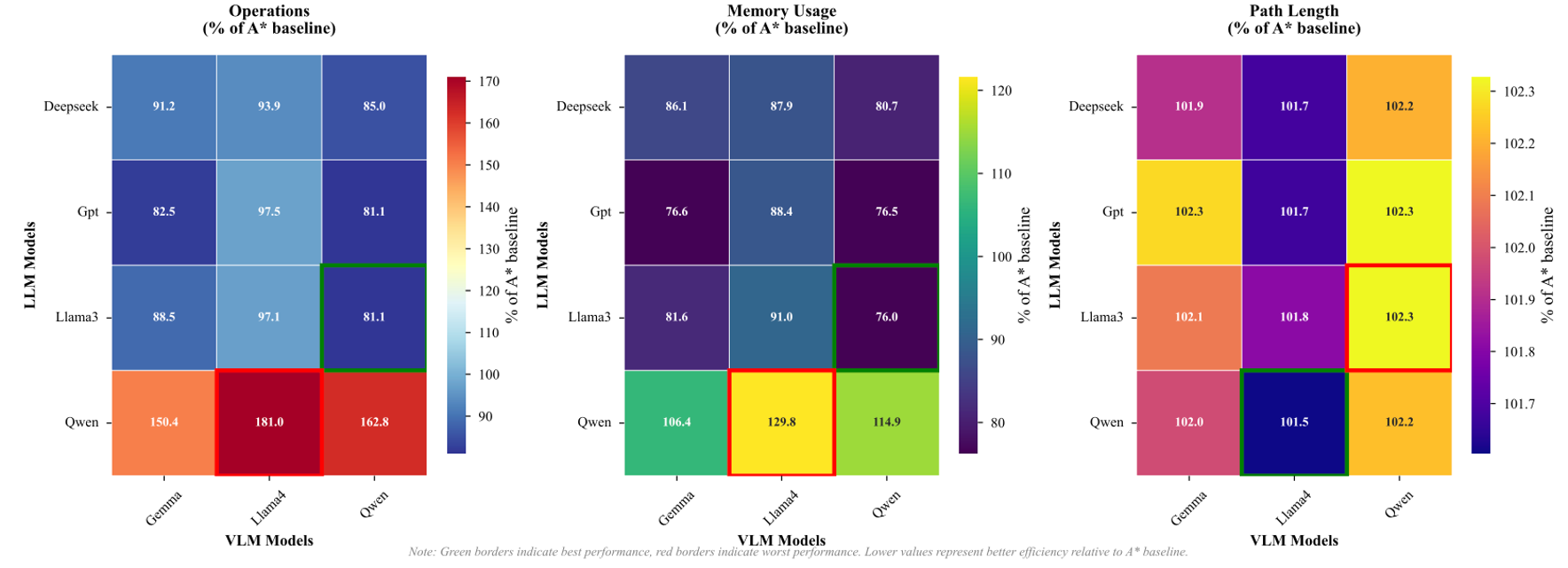}
\caption{
Comparative Efficiency of LLM–VLM Models Across Operations, Memory, and Path Quality
}
\label{Comparative Efficiency}

\end{figure*}

Although \textsc{MMP-A*} consistently enhances efficiency and stability, its overall performance remains sensitive to the intrinsic reasoning capability of the underlying language model. As shown in Figure~\ref{Comparative Efficiency}, configurations using \textsc{Qwen} as the LLM consistently exhibit the weakest results across all metrics, including the highest operation and memory ratios. This is primarily because Qwen often generates misleading or dead-end waypoints that terminate within narrow or blocked corridors. Once these erroneous checkpoints are provided, the accompanying VLM cannot fully recover, the visual module can only validate or prune existing candidates, not generate alternative waypoints, thus propagating early errors through the search process.

This limitation arises not from perceptual failure of the VLM but from the upstream linguistic bias of the LLM, which shapes the initial search manifold. The effectiveness of \textsc{MMP-A*} therefore depends on the LLM’s ability to propose spatially coherent and semantically meaningful waypoint candidates, allowing the VLM to refine rather than rescue the trajectory.

\begin{table*}[!tp]
\centering
\begin{adjustbox}{max width=\textwidth}
\footnotesize
\begin{tabular}{l l l c c c}
\toprule
\textbf{Methodology} & \textbf{Base Model} & \textbf{Prompt Approach} &
\textbf{Operation Ratio $\downarrow$ (\%)} &
\textbf{Storage Ratio $\downarrow$ (\%)} &
\textbf{Relative Path Length $\downarrow$ (\%)} \\
\midrule
\textbf{A*} & -- & -- & 100 & 100 & 100 \\
\midrule
\multirow{12}{*}{\textbf{LLM-A*}}
  & \multirow{3}{*}{DeepSeek-V3}            & Few-Shot & 109.6 & 93.5 & 103.1 \\
  &                                         & CoT      & 131.6 & 106.6 & 102.7 \\
  &                                         & RePE     & 125.2 & 114.6 & 103.4 \\
\cmidrule(lr){2-6}
  & \multirow{3}{*}{Llama-3.3-70B} & Few-Shot & 100.0 & 94.9 & 103.2 \\
  &                                         & CoT      & 110.2 & 102.5 & 102.7 \\
  &                                         & RePE     & 112.1 & 103.1 & 103.3 \\
\cmidrule(lr){2-6}
  & \multirow{3}{*}{Qwen2.5-7B}    & Few-Shot & 173.7 & 133.5 & 103.0 \\
  &                                         & CoT      & 144.4 & 106.3 & 104.0 \\
  &                                         & RePE     & 191.6 & 141.4 & 102.5 \\
\cmidrule(lr){2-6}
  & \multirow{3}{*}{GPT-4o-mini}            & Few-Shot & 127.1 & 119.7 & 103.1 \\
  &                                         & CoT      & 117.7 & 105.1 & 103.2 \\
  &                                         & RePE     & 115.3 & 104.6 & 103.5 \\
\midrule
\multirow{12}{*}{\textbf{MMP-A*}}
  & \multirow{3}{*}{DeepSeek-V3 + Gemma-3n-E4B}            & Few-Shot & 73.0 & 72.6 & 102.2 \\
  &                                                                & CoT      & 79.5 & 77.0 & 102.4 \\
  &                                                                & RePE     & 91.1 & 86.0 & 101.9 \\
\cmidrule(lr){2-6}
  & \multirow{3}{*}{Llama-3.3-70B + Gemma-3n-E4B} & Few-Shot & 72.4 & 71.7 & 102.1 \\
  &                                                                & CoT      & 91.4 & 85.9 & 101.8 \\
  &                                                                & RePE     & 88.5 & 81.5 &  102.0 \\
\cmidrule(lr){2-6}
  & \multirow{3}{*}{Qwen2.5-7B + Gemma-3n-E4B}    & Few-Shot & 137.6 & 102.2 & 102.2 \\
  &                                                                & CoT      & 111.0 & 91.3 & 102.8 \\
  &                                                                & RePE     & 150.4 & 106.3 & 101.9 \\
\cmidrule(lr){2-6}
  & \multirow{3}{*}{GPT-4o-mini + Gemma-3n-E4B}            & Few-Shot & 93.2 & 86.4 & 101.5 \\
  &                                                                & CoT      & 89.5 & 82.5 & 101.6 \\
  &                                                                & RePE     & 82.4 & 76.5 & 102.2 \\
\bottomrule
\end{tabular}
\end{adjustbox}
\caption{Quantitative results of different prompting strategies under \textbf{LLM-A*} and \textbf{MMP-A*}. 
The few-shot, chain-of-thought (CoT), and recursive path evaluation (RePE) settings are compared across multiple base models. }
\label{tab:prompt}
\end{table*}

\begin{table*}[!tb]
\centering
\begin{adjustbox}{max width=\textwidth}
\footnotesize
\begin{tabular}{c c c c c c c c c c c c}
\toprule
\multirow{2}{*}{\textbf{Method}} & \multirow{2}{*}{\textbf{LLM}} & \multirow{2}{*}{\textbf{VLM}} &
\multicolumn{3}{c}{\textbf{Operation Ratio} $\downarrow$} &
\multicolumn{3}{c}{\textbf{Storage Ratio} $\downarrow$} &
\multicolumn{3}{c}{\textbf{Rel.\ Path Length} $\downarrow$} \\
\cmidrule(lr){4-6}\cmidrule(lr){7-9}\cmidrule(lr){10-12}
 &  &  & \textbf{w/ Adap.} & \textbf{w/o Adap.} & $\Delta$ & \textbf{w/ Adap.} & \textbf{w/o Adap.} & $\Delta$ & \textbf{w/ Adap.} & \textbf{w/o Adap.} & $\Delta$ \\
\midrule
\multirow{4}{*}{\textbf{LLM-A*}} 
  & DeepSeek-V3              & -- & 135.8 & 125.2 & $+10.6$ & 118.6 & 114.6 & $+4.0$ & \textbf{\underline{100.9}} & 103.4 & $-2.5$ \\
  & Llama-3.3-70B   & -- & 128.5 & 112.1 & $+16.4$ & 112.9 & 103.1 & $+9.8$ & 101.0 & 103.3 & $-2.3$ \\
  & Qwen2.5-7B      & -- & 182.0 & 191.6 & $\textbf{-9.6}$  & 133.9 & 141.4 & $\textbf{-7.5}$ & 101.2 & 102.5 & $-1.3$ \\
  & GPT-4o-mini              & -- & 136.0 & 115.3 & $+20.7$ & 114.5 & 104.6 & $+9.9$ & \textbf{\underline{100.9}} & 103.5 & $\textbf{-2.6}$ \\
\midrule
\multirow{12}{*}{\textbf{MMP-A*}} 
  & \multirow{3}{*}{DeepSeek-V3}            & Gemma-3n-E4B   & 91.1 & 82.3 & $+8.8$ & 86.0 & 81.0 & $+5.0$ & 101.9 & 103.9 & $-2.0$ \\
  &                                          & Llama 4 Maverick & 93.9 & 84.8 & $+9.1$ & 87.8 & 82.0 & $+5.8$ & 101.6 & 103.9 & $-2.3$ \\
  &                                          & Qwen2.5-VL & 84.9 & 78.7 & $+6.2$ & 80.6 & 76.9 & $+3.7$ & 102.2 & 103.6 & $-1.4$ \\
\cmidrule(lr){2-12}
  & \multirow{3}{*}{Llama-3.3-70B} & Gemma-3n-E4B   & 88.5 & 78.8 & $+9.7$ & 81.5 & 75.9 & $+5.6$ & 102.0 & 103.6 & $-1.6$ \\
  &                                          & Llama 4 Maverick & 97.0 & 86.1 & $+10.9$ & 91.0 & 83.6 & $+7.4$ & 101.8 & 103.6 & $-1.8$ \\
  &                                          & Qwen2.5-VL & 81.0 & 72.2 & $+8.8$ & 76.0 & \textbf{\underline{70.7}} & $+5.3$ & 102.3 & 104.1 & $-1.8$ \\
\cmidrule(lr){2-12}
  & \multirow{3}{*}{Qwen2.5-7B}    & Gemma-3n-E4B   & 150.4 & 141.3 & $+9.1$ & 106.3 & 103.7 & $+2.6$ & 101.9 & 103.4 & $-1.5$ \\
  &                                          & Llama 4 Maverick & 181.0 & 177.4 & $+3.6$ & 129.8 & 128.3 & $+1.5$ & 101.5 & 102.9 & $-1.4$ \\
  &                                          & Qwen2.5-VL & 162.8 & 156.4 & $+6.4$ & 114.8 & 111.9 & $+2.9$ & 102.2 & 103.7 & $-1.5$ \\
\cmidrule(lr){2-12}
  & \multirow{3}{*}{GPT-4o-mini}            & Gemma-3n-E4B   & 82.4 & \textbf{\underline{70.6}} & $+11.8$ & 76.5 & 69.3 & $+7.2$ & 102.2 & 104.4 & $-2.2$ \\
  &                                          & Llama 4 Maverick & 97.4 & 87.0 & $+10.4$ & 88.3 & 82.6 & $+5.7$ & 101.6 & 103.9 & $-2.3$ \\
  &                                          & Qwen2.5-VL & 81.0 & 74.1 & $+6.9$ & 76.4 & 72.5 & $+3.9$ & 102.3 & 103.7 & $-1.4$ \\
\bottomrule
\end{tabular}
\end{adjustbox}
\caption{Comparison of \textsc{LLM-A*} and \textsc{MMP-A*} with and without adaptive decay across LLM--VLM pairs.}

\label{tab:adaptive}
\end{table*}

\begin{figure*}[!tb]
	\centering
	\includegraphics[width=1\linewidth] {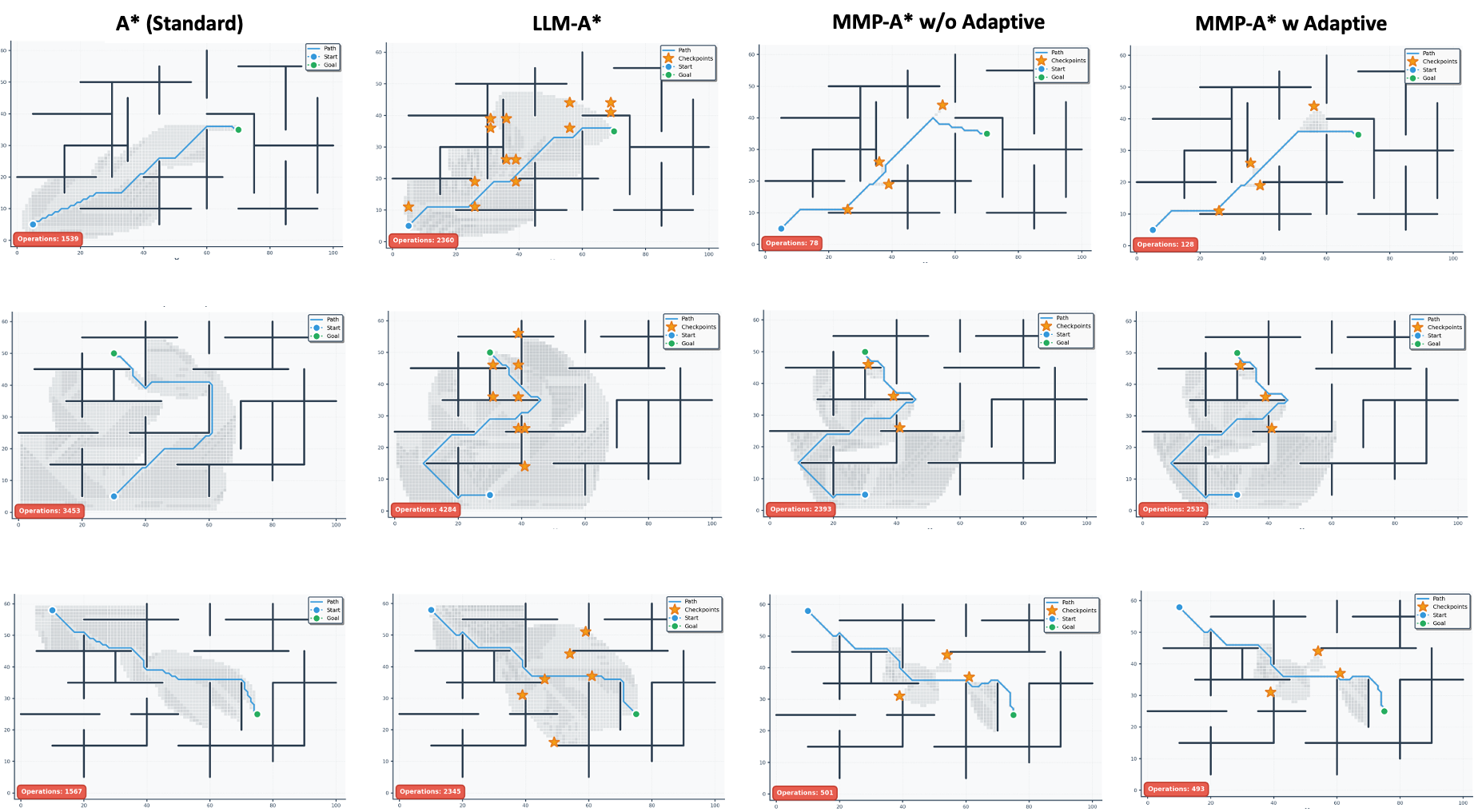}
\caption{
Visual comparison of search behaviors across different planners.
\textsc{LLM-A*} suffers from redundant expansions due to unreliable waypoint guidance, 
while \textbf{MMP-A*} refines these checkpoints through visual filtering and adaptive decay, 
achieving smoother, obstacle-aware trajectories with the fewest search operations and balanced efficiency–optimality trade-off.
}
\label{Visual comparison}
\end{figure*}

\begin{figure*}[!tb]
	\centering
	\includegraphics[width=0.8\linewidth] {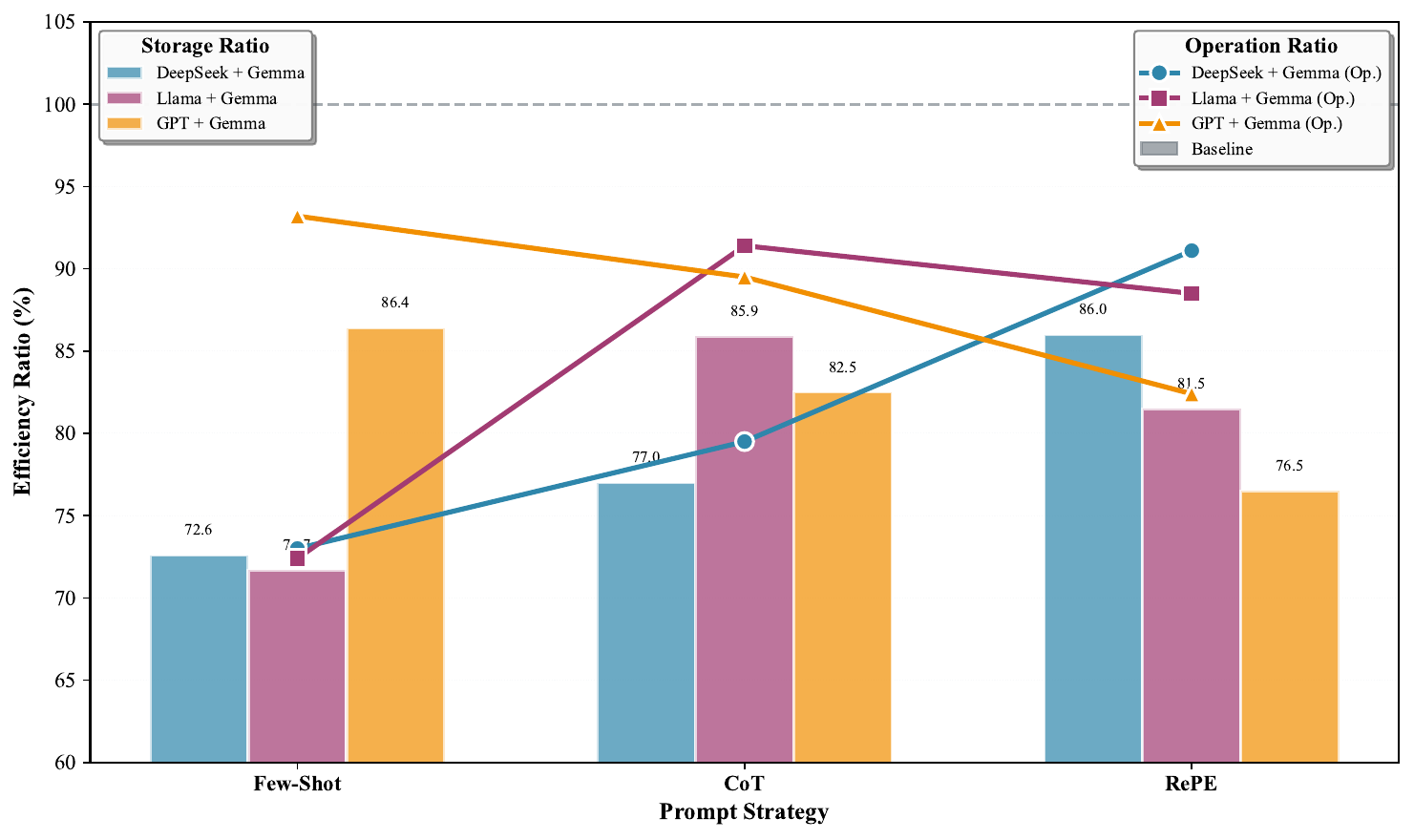}
\caption{
Comparative Efficiency of LLM–VLM Models Across Operations, Memory, and Path Quality
}
\label{prompt_compare}

\end{figure*}

\section{Runtime Performance Trade-Off Analysis}
\label{sec:runtime-trafeoff}

\begin{figure*}[!tb]
	\centering
	\includegraphics[width=1\linewidth] {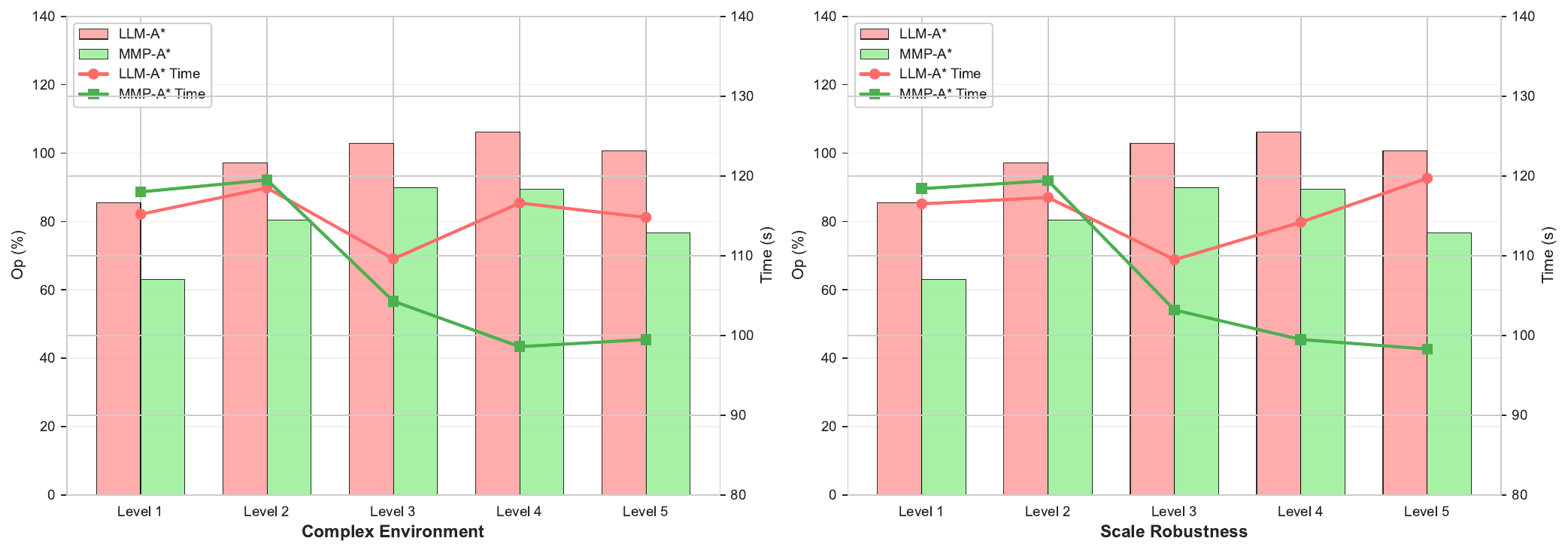}
\caption{
\textbf{Runtime Performance Trade-Off Between LLM-A* and MMP-A*.}
As environment complexity and map resolution increase, MMP-A* not only sustains superior optimality rates but also exhibits a decreasing runtime trend, 
becoming comparable to and even surpassing the runtime of classical A* at the highest complexity and resolution levels.
}
\label{runtime-trafeoff}
\end{figure*}

Figure~\ref{runtime-trafeoff} compares the runtime and operation performance of LLM-A* and MMP-A* across two settings: \textit{Complex Environment} (GPT-4o-mini + Qwen-VL) and \textit{Scale Robustness} (GPT-4o-mini + Gemma). 

As resolution and complexity increase, LLM-A* becomes progressively slower, primarily because it generates redundant or poorly aligned waypoints; these noisy suggestions misguide the search, inflate the A* exploration space, and cause the number of expansions to grow disproportionately with map difficulty. In contrast, \textbf{MMP-A*} exploits structured waypoint reasoning and multimodal filtering to impose a coherent global guidance signal, pruning large portions of the search space and enabling its runtime to converge toward, and at higher difficulty levels even surpass the baseline runtime of classical A*. Although MMP-A* incurs a small overhead on simple, low-resolution maps due to LLM/VLM calls, this cost is effectively amortized in complex and large-scale environments, where guided reasoning sharply compresses the search space. This trend aligns well with real-world robotic navigation, where maps are typically high-resolution and structurally complex, making MMP-A* a practically attractive choice that reconciles multimodal reasoning with stringent runtime constraints.

%% file: tables/prompt.tex
\begin{table*}[tb!]
\centering
\fontsize{10pt}{12pt}\selectfont
\begin{tcolorbox}[colback=purple!8!white, colframe=purple!80!black, 
                  boxrule=1pt, arc=3pt, left=6pt, right=6pt, top=6pt, bottom=6pt, width=0.95\textwidth]
\vspace{0.3cm}

Identify a path between the start and goal points to navigate around obstacles and find the shortest path to the goal. 
Horizontal barriers are represented as [y, x\_start, x\_end], and vertical barriers are represented as [x, y\_start, y\_end].
Conclude your response with the generated path in the format ``Generated Path: [[x1, y1], [x2, y2], ...]''.

\vspace{0.3cm}

Start Point: [5, 5]\\
Goal Point: [20, 20]\\
Horizontal Barriers: [[10, 0, 25], [15, 30, 50]]\\
Vertical Barriers: [[25, 10, 22]]\\
Generated Path: [[5, 5], [26, 9], [25, 23], [20, 20]]

\vspace{0.3cm}
\textcolor{gray}{[5 in-context demonstrations abbreviated]}
\vspace{0.3cm}

Start Point: \{start\}\\
Goal Point: \{goal\}\\
Horizontal Barriers: \{horizontal\_barriers\}\\
Vertical Barriers: \{vertical\_barriers\}\\
Generated Path: \underline{\textbf{Model Generated Answer Goes Here}}
\end{tcolorbox}
\caption{The template of the prompt used for \textsc{MMP-A*} with 5-shot demonstration.}
\label{fig:fewshot}
\end{table*}

\begin{table*}[tb!]
\centering
\fontsize{10pt}{12pt}\selectfont
\begin{tcolorbox}[colback=purple!8!white, colframe=purple!80!black,
                  boxrule=1pt, arc=3pt, width=0.95\textwidth,
                  left=6pt, right=6pt, top=6pt, bottom=6pt]

\vspace{0.3cm}

Identify a path between the start and goal points to navigate around obstacles and find the shortest path to the goal.  
Horizontal barriers are represented as [y, x\_start, x\_end], and vertical barriers are represented as [x, y\_start, y\_end].  
Conclude your response with the generated path in the format ``Generated Path: [[x1, y1], [x2, y2], ...]''.

\vspace{0.3cm}

Start Point: [5, 5]\\
Goal Point: [20, 20]\\
Horizontal Barriers: [[10, 0, 25], [15, 30, 50]]\\
Vertical Barriers: [[25, 10, 22]]\\
Thought: Identify a path from [5, 5] to [20, 20] while avoiding the horizontal barrier at y=10 spanning x=0 to x=25 by moving upwards and right, then bypass the vertical barrier at x=25 spanning y=10 to y=22, and finally move directly to [20, 20].\\
Generated Path: [[5, 5], [26, 9], [25, 23], [20, 20]]

\vspace{0.3cm}
\textcolor{gray}{[3 in-context demonstrations abbreviated]}
\vspace{0.3cm}

Start Point: \{start\}\\
Goal Point: \{goal\}\\
Horizontal Barriers: \{horizontal\_barriers\}\\
Vertical Barriers: \{vertical\_barriers\}\\
Generated Path: \underline{\textbf{Model Generated Answer Goes Here}}

\end{tcolorbox}
\caption{Prompt template for \textsc{MMP-A*} using 3-shot Chain-of-Thought (CoT) reasoning.}
\label{fig:cot}
\end{table*}

\begin{table*}[tb!]
\centering
\fontsize{10pt}{12pt}\selectfont
\begin{tcolorbox}[colback=purple!8!white, colframe=purple!80!black,
                  boxrule=1pt, arc=3pt, width=0.95\textwidth,
                  left=6pt, right=6pt, top=6pt, bottom=6pt]

\vspace{0.3cm}

Identify a path between the start and goal points to navigate around obstacles and find the shortest path to the goal.  
Horizontal barriers are represented as [y, x\_start, x\_end], and vertical barriers are represented as [x, y\_start, y\_end].  
Conclude your response with the generated path in the format ``Generated Path: [[x1, y1], [x2, y2], ...]''.

\vspace{0.3cm}

Start Point: [5, 5]\\
Goal Point: [20, 20]\\
Horizontal Barriers: [[10, 0, 25], [15, 30, 50]]\\
Vertical Barriers: [[25, 10, 22]]\\
-- First Iteration on [5, 5]\\
Thought: The horizontal barrier at y=10 spanning x=0 to x=25 blocks the direct path. Move to the upper-right corner of the barrier.\\
Selected Point: [26, 9]\\
Evaluation: The point [26, 9] bypasses the horizontal barrier efficiently.\\
-- Second Iteration on [26, 9]\\
Thought: The vertical barrier at x=25 blocks direct motion to [20, 20]; move around it.\\
Selected Point: [25, 23]\\
Evaluation: The new point successfully avoids the barrier.\\
-- Third Iteration on [25, 23]\\
Thought: No further obstacles to the goal.\\
Selected Point: [20, 20]\\
Generated Path: [[5, 5], [26, 9], [25, 23], [20, 20]]

\vspace{0.3cm}
\textcolor{gray}{[3 in-context demonstrations abbreviated]}
\vspace{0.3cm}

Start Point: \{start\}\\
Goal Point: \{goal\}\\
Horizontal Barriers: \{horizontal\_barriers\}\\
Vertical Barriers: \{vertical\_barriers\}\\
Generated Path: \underline{\textbf{Model Generated Answer Goes Here}}

\end{tcolorbox}
\caption{Prompt template for \textsc{MMP-A*} using 3-shot Recursive Path Evaluation (RePE) reasoning.}
\label{fig:repe}
\end{table*}

\begin{table*}[tb!]
\centering
\fontsize{10pt}{12pt}\selectfont
\begin{tcolorbox}[colback=green!5!white, colframe=green!60!black,
                  boxrule=1pt, arc=3pt, width=0.95\textwidth,
                  left=6pt, right=6pt, top=6pt, bottom=6pt]
\vspace{0.3cm}

You are presented with two visual representations of the same maze environment. The obstacles are defined by two distinct visual cues: standard black grid walls and irregular red regions delineated by dashed contours: \\
1. First image: Shows the clean map with start point (blue square) and goal point (green square). \\
2. Second image: Shows the same map with num-waypoints waypoints (yellow stars) placed along a blue path. Waypoints are indexed 1..{num-waypoints}; goal is id {num-waypoints + 1}.
\\
What is a "waypoint" and its role (read carefully):
\begin{itemize}
    \item A waypoint (yellow star) is a navigation landmark, a coarse checkpoint placed in clearly open space that helps the robot orient its heading and follow a feasible route.
    \item It is NOT a precise docking coordinate. Waypoints indicate: 
    \begin{itemize}
        \item Turning points (where the robot must change direction)
        \item Corridor transitions (entering or leaving a corridor)
        \item Decision junctions (where multiple passages meet)
    \end{itemize}
    \item A valid waypoint MUST be centered in open space with visible clearance from walls. Waypoints in dead-ends, touching/near walls, or inside narrow squeezes are invalid and must be discarded.
    \item Because the robot travels in straight-line segments between consecutive waypoints, every such segment in the final path must be visibly open and free of contact with barriers.
    \item \textbf{Important:} The second image (with yellow stars) is only \textbf{a suggested route}, it is NOT guaranteed to be a valid robot path. You must infer safety from the barrier layout (do not assume the blue path is correct).
\end{itemize}

\textbf{IMPORTANT RULES:}
\begin{itemize}
    \item This is for a physical robot. The robot cannot touch, graze, or squeeze between walls. Be conservative: if a straight segment is ambiguous or appears to touch walls, treat it as blocked.
    \item Do NOT create any new waypoints. Choose only from the existing numbered candidate waypoints shown in the second image. Do NOT output internal chain-of-thought. Output only the structured JSON described below using factual, image-tied statements.
\end{itemize}

\textbf{TASK (two stages, output combined):}
\begin{enumerate}
    \item First, inspect the clean map (first image) globally and identify which corridors or directions from start toward goal are visibly open or blocked.
    \item Then, using that global view, evaluate each original waypoint in order and decide whether it is essential as a navigation marker:
    \begin{itemize}
        \item Keep a waypoint if it lies in open space and is necessary as a turning point, corridor transition, or decision marker so that start → selected-waypoint-1 → ... → goal can be realized by clearly open straight segments.
        \item Discard a waypoint if it lies in a blocked, narrow, redundant, or dead-end location that would force the robot into unsafe or blocked segments.
    \end{itemize}
\end{enumerate}

\textbf{OUTPUT (strict JSON only; nothing else):}
\begin{itemize}
    \item  \textbf{``selected-waypoints":} [ list of integer waypoint IDs to KEEP in traversal order, e.g. [2, 5] ], must contain only integers between 1 and {num-waypoints}. If no original waypoint is needed, return an empty list.
    \item  \textbf{``final-reasoning":} ``(a) explicitly describe the overall feasible route(s) observed on the clean map before considering waypoints, and (b) explain for each chosen waypoint why it is necessary and for discarded waypoints why they were removed. Keep statements factual and tied to visible barriers/corridors, must be factual and image-referential"
\end{itemize}


\end{tcolorbox}
\caption{Prompt template used in \textsc{MMP-A*} for filtering spatially valid waypoints from paired maze images.}
\label{fig:vlm_template}
\end{table*}